%% file: main.tex
\documentclass{article}

\usepackage{arxiv}

\usepackage[utf8]{inputenc} 
\usepackage[T1]{fontenc}    
\usepackage{hyperref}       
\usepackage{url}            
\usepackage{booktabs}       
\usepackage{amsfonts}       
\usepackage{nicefrac}       
\usepackage{microtype}      
\usepackage{lipsum}		
\usepackage{graphicx}
\usepackage{natbib}
\usepackage{doi}
\input{math_commands.tex}

\usepackage{hyperref}
\usepackage{times}
\usepackage{graphicx}
\usepackage{dsfont}
\usepackage{mymacros}

\newcommand{\note}[1]{
}
\newcommand{\indep}{\perp \!\!\! \perp}
\newcommand{\notindep}{\not\!\perp\!\!\!\perp}

\title{Localizing predictive concepts encoded in neural representations via hierarchical testing}

\date{} 					

\newcommand{\orcid}[1]{\href{https://orcid.org/#1}{\includegraphics[scale=0.06]{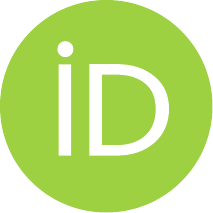}}}
\newcommand{\authorblock}{%
\parbox{0.95\textwidth}{\centering\normalfont
{\bfseries\large
\orcid{0009-0009-1691-4919}\hspace{1mm}Viktoriia Huryn\textsuperscript{1,2*},\quad
\orcid{0009-0004-9818-852X}\hspace{1mm}Marco Simnacher\textsuperscript{2,*},\quad
\orcid{0009-0006-8259-7278}\hspace{1mm}Jim Berend\textsuperscript{3},\quad
\orcid{0000-0002-6283-3265}\hspace{1mm}Wojciech Samek\textsuperscript{3,4,5},\quad
\orcid{0000-0001-6363-2556}\hspace{1mm}Christoph Lippert\textsuperscript{6},\quad
\orcid{0000-0003-0495-850X}\hspace{1mm}Sonja Greven\textsuperscript{2, \dag},\quad
\orcid{0000-0002-0881-3116}\hspace{1mm}Uwe Ohler\textsuperscript{1,2,4 \dag}\par}
\vspace{1em}

{\small\itshape\normalfont
\textsuperscript{1}Max Delbr\"uck Center for Molecular Medicine, Berlin, Germany\par
\textsuperscript{2}Humboldt-Universit\"at zu Berlin, Germany\par
\textsuperscript{3}Department of Artificial Intelligence, Fraunhofer Heinrich Hertz Institute, Berlin, Germany\par
\textsuperscript{4}BIFOLD -- Berlin Institute for the Foundations of Learning and Data, Germany\par
\textsuperscript{5}Department of Electrical Engineering and Computer Science, Technical University of Berlin, Germany\par
\textsuperscript{6}Hasso Plattner Institute for Digital Engineering, Germany\par}
\vspace{0.8em}

{\small
\textsuperscript{*}These authors contributed equally. \\
\textsuperscript{\dag}Corresponding authors: \texttt{sonja.greven@hu-berlin.de}, \texttt{uwe.ohler@mdc-berlin.de}\par}
}%
}

\author{\authorblock}

\renewcommand{\headeright}{}
\renewcommand{\undertitle}{}
\renewcommand{\shorttitle}{Localizing predictive concepts encoded in neural representations via hierarchical testing}

\hypersetup{
pdftitle={Localizing predictive concepts encoded in neural representations via hierarchical testing},
pdfsubject={},
pdfauthor={V Huryn, M Simnacher, et al.},
pdfkeywords={},
}

\begin{document}
\maketitle

\begin{abstract}
Understanding which regulatory sequence motifs a genomic deep learning model has learned, with statistical error guarantees, is essential for trusting its use in downstream tasks such as variant effect prediction.
Existing approaches to identifying which units in a trained network encode human-understandable concepts fall into two groups. Exploratory methods rank and visualize units but offer no statistical guarantee that the selected units are associated with the prediction target across the population of inputs. Testing-based methods test only single units or one pre-defined set of them, and do not control the error across the many sets  of units. 
We introduce CUTTIE (Cluster-tree Unit Top-down Testing of Independence with Error control), which clusters the units of each layer into a tree and tests the clusters top-down with conditional independence tests, localizing information about the target while controlling the family-wise error rate. Thereby, we identify clusters that carry information about the target beyond the rest of the layer.
For the units in the identified clusters, conditional attribution maps then reveal what motifs they encode.
In a controlled simulation, the procedure identifies the clusters whose concepts affect the target. 
We apply the procedure to a convolutional model predicting cell-type-specific chromatin accessibility from DNA sequence, where the concepts encoded in the identified clusters map onto known regulatory motifs.
\end{abstract}


\section{Introduction}
Concept-based explanations go beyond input attribution by relating a model’s internal representations to human-understandable concepts, either user-defined or discovered automatically, and by quantifying how much these concepts influence the model’s predictions.
Deep neural networks that predict molecular readouts such as chromatin accessibility from DNA sequence are widely used to estimate variant effects, i.e., how a mutation changes the predicted activity. Trusting these estimates requires checking that the change is driven by the correct regulatory concepts, such as transcription factor binding motifs (short DNA patterns recognized by regulatory proteins), which in turn requires knowing how and where in the network such concepts are encoded.
In computer vision, attribution maps of individual units can often be interpreted directly by visual inspection. In genomic models that predict DNA activity from sequence, however, attribution maps cannot be interpreted with biological concepts by visual inspection alone and require downstream analyses such as motif discovery. This makes it essential to narrow down which units are worth analyzing.

Most existing approaches to identifying these units are exploratory. Methods such as TCAV \citep{TCAV}, ACE \citep{ACE}, and CRAFT \citep{CRAFT} surface attribution maps or concept-activation scores that a human judges plausible, with no statement of statistical confidence. 
However, a unit can appear highly relevant to a prediction by chance alone. Establishing that it is genuinely connected to the prediction target therefore requires population-level evidence with explicit error control.
A smaller line of work adds formal hypothesis testing, but each instance addresses only part of the problem. \citet{bykov2023labeling} test whether a unit's activations differ between concept-positive and concept-negative samples, but treat units independently and apply no multiple-testing correction (p-values are evaluated at a nominal 0.05). \citet{teneggi2024testing} test conditional independence between model predictions and a concept and do correct for multiple testing, but only across a small set of pre-specified concepts. Neither addresses which units or clusters of units in a layer carry predictive information, while controlling the error across the many candidates. 
Such error control across nested clusters of correlated variables is provided by hierarchical testing procedures \citep{meinshausen2008hierarchical, mandozzi2016hierarchical}, but they were formulated for tests in linear models and have not been applied to network units, which are strongly correlated, come without a given hierarchy, and lack established hypotheses and test statistics.
Since their error control only requires a hierarchy fixed in advance and a valid p-value for each cluster \citep{goeman2010sequential}, these tests can be replaced, as \citet{paillard2025hierarchical} do with conditional permutation importance for the inputs of a prediction model.
We instead test units inside a network and allow any conditional independence test (CIT), a general class that includes such importance-based tests. We use the generalized covariance measure \citep[GCM;][]{shah2020hardness}, which only requires flexible regressions rather than conditional sampling.
Such tests have been successfully applied to fixed network representations \citep{simnacher2026deep, simnacher2026embedded}, whose features are strongly correlated and depend nonlinearly on each other, but only to test a representation as a whole.
Other hierarchical approaches for network inputs cluster them by their contribution to the network's prediction, but without error control \citep[e.g.,][]{ACD_new}.

We propose CUTTIE (Cluster-tree Unit Top-down Testing of Independence with Error control) -- a hierarchical testing procedure for localizing information about the prediction target within the layers of a trained network while controlling the family-wise error rate (FWER).
Per layer, we cluster units into a tree based on their scores, such as activations or relevance scores. 
Traversing the tree top-down, we apply conditional independence tests (CITs) to the target and the scores of the units in each cluster given the scores of the remaining units, and thereby identify which clusters
carry information about the target beyond the rest of the layer. 
The procedure applies to activations, which identify units that encode predictive information, and to relevance scores from Concept Relevance Propagation (CRP) \citep{CRP}, which identify units the model relies on for its predictions. Conditional attribution maps of the selected units then show which concepts they encode. To our knowledge, this is the first framework to combine formal, hierarchical error-rate control with concept-level attribution in neural networks.

We demonstrate CUTTIE in a challenging genomics setting, where concepts cannot be interpreted directly from visualizations, some motifs are universally active across cell types and can be spuriously associated with cell-type-specific outcomes, experimental noise is high, and only a limited number of units and their encoded motifs can be validated against a motif database. We apply CUTTIE to a sequence-to-activity (S2A) model trained to predict DNA activity from sequence across cell types of the zebrafish embryo. CUTTIE identifies non-redundant units that are predictive of DNA activity in individual cell types. Downstream analysis shows that these units correspond to known regulatory motifs, and that the model relies on them to distinguish between cell types: the motifs recovered from their attribution maps agree with the cell types in which they are reported to be active in the literature.

Our contributions are: (1) a hierarchical testing procedure that localizes predictive information in neural network layers with FWER control, for activation and CRP scores, together with a characterization of the hypotheses tested at each node for a fixed, trained network; (2) conditional independence tests suited to vector-valued responses and the strong correlations among units; and (3) a simulation with known ground truth and an application to a DNA sequence-to-activity model.

\section{Method}
\label{sec:hierarchical-testing}

We consider a neural network $f_\theta$ fitted to predict $Y$ from $X$, where $Y \in \mathbb R^K$ is a vector of $K$ responses and $X \in \mathcal X$. In our application, $X \in \{0,1\}^{4\times p}$ is a one-hot encoded DNA sequence of length $p$, and $Y$ contains its responses in $K$ cell types, which we refer to as classes. 
We formulate the method for this setting, i.e.\ for information about differences between the class responses. Since the procedure only requires a fixed network, unit scores and a valid test at each node, it applies to any input space $\mathcal X$ and, without centering (Subsection~\ref{subsec:hypotheses}), to responses that are not compared across classes, including $K=1$. 
Fixing the fitted parameters $\hat\theta$, we aim to find units and clusters of units of a layer whose scores carry information about these differences beyond the remaining units, with FWER control for a distribution $P_{XY}$ of $(X,Y)$. 
To this end, we compute scores for each unit (Subsection~\ref{subsec:unit_scores}), state the hypothesis for a cluster of units (Subsection~\ref{subsec:hypotheses}), and organize the units of the layer in a tree of nested clusters, fixed before any p-value is computed (Subsection~\ref{subsec:clustering}). A hierarchical procedure tests the nodes top-down, descending to a cluster's children only if its hypothesis is rejected (Subsection~\ref{subsec:hierarchical_testing}), which controls the FWER at level $\alpha$ if each node test gives a valid p-value (Subsection~\ref{subsec:tests}). Subsections~\ref{subsec:localizing_classes}--\ref{sec:attribution-maps} localize the classes, interpret rejections and visualize the patterns detected by the units in rejected clusters.

\begin{figure*}[t]
  \centering
  \includegraphics[width=\textwidth]{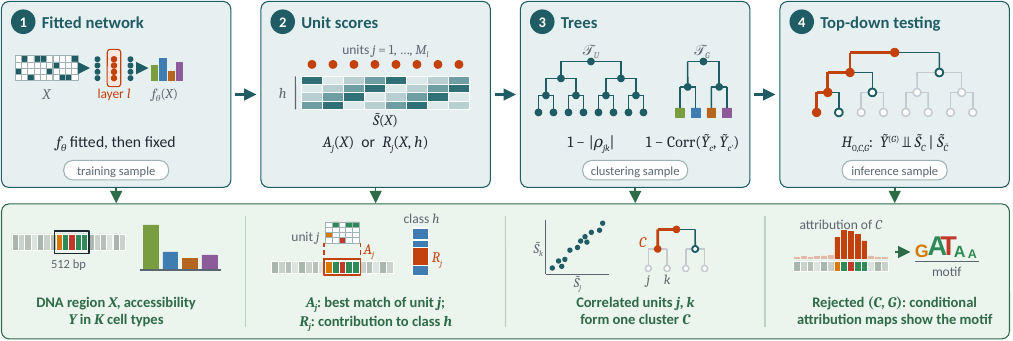}
  \caption{
  Overview of CUTTIE. The proposed procedure (top) and its reading in the application (bottom).
   (1)~The network $f_\theta$ is fitted on a training sample and then fixed.
(2)~Each unit $j$ of layer $l$ receives as scores its activation $A_j(X)$ ($H=1$) or its CRP relevance $R_j(X,h)$ for each class $h$ ($H=K$), collected in $\tilde S(X)$.
(3)~On a clustering sample, complete linkage builds a tree $\mathcal T_U$ over the units, in which units with strongly correlated scores form one cluster $C$, and a tree $\mathcal T_G$ whose nodes $G$ are clusters of classes.
(4)~On an inference sample, $\mathcal{T}_U$ is tested top-down for each internal node $G$ of $\mathcal{T}_G$ (orange: rejected), with FWER control at level $\alpha$ (Figure~\ref{fig:node_tests}). The hypothesis $H_{0,C,G}$, shown for activations, states that the units in $C$ carry no information about differences between the classes in $G$ beyond the rest of the layer (Subsection~\ref{subsec:localizing_classes}). For CRP scores, with responses and scores centered within $G$, it reads $\tilde Y^{(G)}_c \indep \tilde S^{(G)}_{cC} \mid \tilde S^{(G)}_{c\bar C}$ for all $c\in G$, where the index $c$ selects the response and the scores for class $c$.
For a rejected pair $(C,G)$, conditional attribution maps show which input patterns, here motifs, the units in $C$ detect.
}
  \label{fig:procedure}
\end{figure*}

The procedure is illustrated in Figure~\ref{fig:procedure}.

\subsection{Unit scores}\label{subsec:unit_scores}

For a fixed layer \(l\) of width \(M_l\), each unit \(j\in\{1,\ldots,M_l\}\) receives \(H\) scalar scores \(S_{hj}(X)\), \(h\in\{1,\ldots,H\}\), per input, where each \(S_{hj}\) is a fixed function of the fitted network \(f_{\hat\theta}\). The number \(H\) depends only on the type of score: \(H=1\) for class-independent scores such as activations, and \(H=K\) for class-specific scores such as CRP scores, where \(h\) indexes the class 
that is scored.
We consider two scores:
\begin{align*}
    S_{hj}(X)=
    \begin{cases}
    A_j(X), & H=1 \quad \text{(activation over sequence)},\\
    R_j(X,h), & H=K \quad \text{(CRP relevance class \(h\))}.
    \end{cases}
\end{align*}
For activations, let $A_{jt}(X)$ denote the activation of unit $j$ at sequence position $t$, and let $A_j(X) = \max_t A_{jt}(X)$ be its strongest response to any window of the input. This score is defined independently of the pooling used in the network. Activations do not depend on the class, so each unit has a single score, and $H=1$. 
For CRP scores, we initialize the relevance at the output \(f_{\hat\theta}(X)_h\), redistribute it to the units of layer \(l\), and sum the resulting relevance map of unit \(j\) over the sequence to obtain \(R_j(X,h)\). Each input is attributed separately for every class \(h\in\{1,\ldots,K\}\), so each unit has one score per class. Activations characterize what a unit detects in the input, whereas CRP scores additionally reflect how the downstream network uses the unit for each class.

For \(H=K\), we center the scores of each unit across classes, \(\tilde S_{hj}(X)=S_{hj}(X)-K^{-1}\sum_{h'=1}^K S_{h'j}(X)\), which removes the component shared by all classes, as it carries no information about differences between them.
For $H = 1$, we do not center, omit the row index, and write $\tilde S_j(X) = S_{1j}(X)$; when the scores are constant across classes, we also write $\tilde S_{cj}(X) = \tilde S_j(X)$ for all classes $c$.
We collect the centered scores of layer \(l\) in the matrix \(\tilde S=(\tilde S_{hj}(X))_{h,j}\in\mathbb{R}^{H\times M_l}\) and omit the argument \(X\) when it is clear from the context. For a cluster of units \(C\subseteq\{1,\dots,M_l\}\) with complement \(\bar C=\{1,\dots,M_l\}\setminus C\), we write \(\tilde S_C\) and \(\tilde S_{\bar C}\) for the columns of \(\tilde S\) belonging to the units in \(C\) and \(\bar C\), and \(\tilde S_{cC}\) and \(\tilde S_{c\bar C}\) for their entries for class \(c\).

\subsection{Hypotheses}\label{subsec:hypotheses}

To detect units whose scores carry information about differences between the class responses, we center the responses across classes, $\tilde Y_c=Y_c-K^{-1}\sum_{c'=1}^K Y_{c'}$, which removes the component shared by all classes. We write $\tilde Y=(\tilde Y_1,\ldots,\tilde Y_{K})^\top\in\mathbb{R}^{K}$.
If differences between the responses are not of interest, e.g.\ for \(K=1\), we do not center, i.e.\ then \(\tilde Y=Y\) and \(\tilde S_{hj}=S_{hj}\).

For scores that do not depend on the class (\(H=1\)), such as activations, we test
\begin{align}\label{equ:hypothesis_ci}
H_{0,C}:\ \tilde Y\indep \tilde S_C \mid \tilde S_{\bar C} \;\; \text{vs.} \;\; H_{1,C}:\ \tilde Y\notindep \tilde S_C \mid \tilde S_{\bar C}.
\end{align}

For scores that depend on the class (\(H=K\)), such as CRP scores, we test
\begin{equation}\label{eq:H0crp}
\begin{aligned}
  H_{0,C}&:\ \tilde Y_c \indep \tilde S_{cC} \mid \tilde S_{c\bar C} \ \text{ for all } c\in\{1,\dots,K\} \quad \text{vs.}\\
  H_{1,C}&:\ \tilde Y_c \notindep \tilde S_{cC} \mid \tilde S_{c\bar C} \ \text{ for some } c\in\{1,\dots,K\}.
\end{aligned}
\end{equation}
which relates the scores for each class only to the response of the same class.

\subsection{Hierarchy and localization}\label{subsec:hierarchical_testing}

To localize clusters and units whose scores carry information about the response differences, we apply the hierarchical testing procedure of \citet{meinshausen2008hierarchical, mandozzi2016hierarchical} separately to each layer.
For layer $l$, we first remove units whose scores are constant on the clustering sample (see below), e.g.\ filters that never activate or units with zero relevance for all classes.
Such units cannot be clustered (Subsection~\ref{subsec:clustering}), and the hypotheses then refer to the remaining units, whose number we denote by $M_l$ from here on.
We then construct one binary tree $\mathcal{T}_U$ whose nodes $C\subseteq\{1,\ldots,M_l\}$ are clusters of units: the root is the whole layer, the leaves are single units, and any two nodes are either disjoint or one contains the other (cf.\ Figures~\ref{fig:procedure} and~\ref{fig:node_tests}).
At each node $C$, we test \eqref{equ:hypothesis_ci} for activations and \eqref{eq:H0crp} for CRP scores, and denote its p-value by $p_C$.
We traverse $\mathcal{T}_U$ top-down and reject $H_{0,C}$ iff the parent of $C$ was rejected (or $C$ is the root) and
\[
p_C^{\mathrm{adj}}=\min\bigl(1,\; p_C\, M_l/|C|\bigr)\le\alpha .
\]
Hence, the root is tested at level $\alpha$ and a single unit at level $\alpha/M_l$, with clusters in between. We call $C$ null if $H_{0,C}$ holds and non-null otherwise. A false rejection is the rejection of a null $C$. The FWER over the tree $\mathcal T_U$ is the probability of at least one false rejection. The procedure controls the FWER at level $\alpha$ if the FWER is at most $\alpha$ whichever clusters are null.

\citet{meinshausen2008hierarchical} and \citet{mandozzi2016hierarchical} formulate the procedure for the coefficients of a linear model, but their proof of FWER control only uses that the tree is fixed in advance (Subsection~\ref{subsec:clustering}) and that each p-value is valid, i.e.\ $P(p_C\le u)\le u$ for all $u\in[0,1]$ if $H_{0,C}$ holds \citep[see also Appendix~\ref{app:fwer} and ][]{goeman2010sequential}.
The FWER is therefore controlled for any hypotheses $H_{0,C}$ with valid p-values, and asymptotically for asymptotically valid ones, so we can replace the linear model by conditional independence tests (CITs), which do not assume a linear relation between scores and responses.

In addition to the training sample of $f_{\hat\theta}$, we use two further samples.
On a clustering sample, we construct $\mathcal{T}_U$ (Subsection~\ref{subsec:clustering}).
On an inference sample of i.i.d.\ copies $(X_1, Y_1), \ldots, (X_n, Y_n)$ of $(X, Y)$, we compute the p-values $p_C$ (Subsection~\ref{subsec:tests}).
For the $i$-th pair, we write $S_i$, $\tilde S_i$ and $\tilde Y_i$ for the scores, centered scores and centered responses, and $\tilde S_{iC}$, $\tilde S_{i\bar C}$ for the columns of $\tilde S_i$ belonging to $C$ and $\bar C$.
Only the inference sample has to be drawn from $P_{XY}$; the training and clustering samples may come from any distribution, as long as they are independent of the inference sample.
Given these two samples, the model and $\mathcal{T}_U$ are fixed, and FWER control holds conditionally on them.

\subsection{Tree construction}\label{subsec:clustering}

Any tree constructed independently of the inference sample yields FWER control (Subsection~\ref{subsec:hierarchical_testing}), so the clustering only affects power and interpretation, and may, in general, be based on different unit scores than the tests.

We construct \(\mathcal{T}_U\) on a clustering sample by complete-linkage clustering with distance \(1-|\rho_{jk}|\) between units \(j\) and \(k\), where \(\rho_{jk}\) is a correlation of their scores on the clustering sample. Complete linkage ensures that clusters low in the tree contain only mutually strongly correlated units, which are largely redundant with respect to their scores. Such units are hard to detect individually by conditional tests, whereas the cluster as a whole can still be detected \citep{mandozzi2016hierarchical}.

For activation scores (\(H=1\)), \(\rho_{jk}=\mathrm{Corr}(\tilde S_j,\tilde S_k)\) is the correlation of the two units' activations across the clustering sample inputs.
For CRP scores ($H = K$), $\rho_{jk}$ is the pooled within-class correlation of the scores (Appendix~\ref{app:crp_corr}). Since the tests regress separately for each class (Subsection~\ref{subsec:tests}), units are redundant only if their scores co-vary across inputs within a class.

\subsection{Tests at each node}\label{subsec:tests}

If $(\tilde Y, \tilde S)$ has an absolutely continuous distribution, no test that controls the type I error uniformly over the null has power greater than the level of the test against any alternative \citep{shah2020hardness}. CITs with power therefore rely on assumptions that restrict the null hypothesis. We use the generalized covariance measure (GCM) of \citet{shah2020hardness}, which is applicable to multivariate responses and scores. It regresses the responses and the scores of the units in $C$ on the scores of the remaining units, and tests whether the residuals are correlated. Its validity requires that these regressions estimate the conditional means sufficiently well, and it has power only if $\mathbb E[\operatorname{Cov}(\tilde Y_c, \tilde S_{cj} \mid \tilde S_{c\bar C})] \neq 0$ for some class $c$ and unit $j \in C$, i.e. if the residuals are correlated in expectation. 

\paragraph{Residual products.}
For each class $c$, the responses and the scores of the units in $C$ are regressed on the class-$c$ scores $\tilde S_{c\bar C}$ of the remaining units. For activations, these are the activations of the remaining units, which are the same for every class, i.e.\ $\tilde S_{cj} = \tilde S_j$ for all $c$. For CRP scores, \eqref{eq:H0crp} relates the response of each class only to the relevance for the same class, given the relevance of the remaining units for that class. We therefore regress only on the relevances $\tilde S_{c\bar C}$ for class $c$. Any regression method can be used, e.g.\ post-lasso (Appendix~\ref{app:implementation}). At the root, $\bar C = \emptyset$, and the regressions reduce to sample means.
Let $\hat m^Y_c(\tilde S_{c\bar C})$ and $\hat m^S_{cj}(\tilde S_{c\bar C})$ be estimates of $E[\tilde Y_c \mid \tilde S_{c\bar C}]$ and $E[\tilde S_{cj} \mid \tilde S_{c\bar C}]$. For each class $c = 1,\dots,K$ and unit $j \in C$, we compute the residual product
\[
  T_{icj} = \big(\tilde Y_{ic} - \hat m^Y_c(\tilde S_{ic\bar C})\big)
            \big(\tilde S_{icj} - \hat m^S_{cj}(\tilde S_{ic\bar C})\big),
\]
and collect these in $T_i \in \mathbb R^{K|C|}$. For activations, $\hat m^S_{cj}(\tilde S_{c\bar C})$ does not depend on $c$.

\paragraph{Test statistic.}
Let $\bar T = n^{-1}\sum_{i=1}^n T_i$ with entries $\bar T_{cj}$, let $\hat\sigma^2_{cj}$ be the empirical variance of $T_{1cj},\dots,T_{ncj}$, and let $\hat\Gamma$ be the empirical correlation matrix of $T_1,\dots,T_n$. Entries with $\hat\sigma_{cj} = 0$ are removed. Under $H_{0,C}$, the standardized entries $\sqrt n\,\bar T_{cj}/\hat\sigma_{cj}$ are asymptotically jointly normal with mean zero and covariance $\Gamma$, the correlation matrix of $T_i$ \citep{shah2020hardness}. 
We aggregate the standardized entries over classes $c\in \{1,\ldots,K\}$ and units $j \in C$ by
\[
W^{\max}_C = \max_{c\in \{1,\ldots,K\},\,j \in C}
\frac{\sqrt n\,|\bar T_{cj}|}{\hat\sigma_{cj}},
\]
and obtain its p-value $p_C$ by comparing $W^{\max}_C$ with the maximum absolute entry of $B$ independent draws from $N(0,\hat\Gamma)$ \citep[Section~3.2]{shah2020hardness}, cf. Appendix~\ref{app:pvalue}.
$W^{\max}_C$ has most power if the signal is concentrated in few entries, i.e.\ few units and classes; an alternative quadratic statistic with more power against signals spread over many entries is given in Appendix~\ref{app:tests}.

\subsection{Localizing classes}\label{subsec:localizing_classes}

Our procedure identifies clusters whose scores carry information about any difference between the class responses. To identify which classes this information corresponds to, we additionally construct a binary tree $\mathcal T_G$ whose nodes $G\subseteq\{1,\dots,K\}$ are clusters of classes, by complete-linkage clustering on the clustering sample with distance $1-\mathrm{Corr}(\tilde Y_c,\tilde Y_{c'})$ so that classes whose responses vary together are grouped. The leaves of $\mathcal T_G$ are single classes, for which the centered responses below vanish, so we test only at the $K-1$ internal nodes, the root and $K-2$ others, and $G$ denotes an internal node in the following. For each $G$ and $c \in G$, we center the responses and, for CRP scores, the scores within $G$,
$\tilde Y^{(G)}_c = Y_c - |G|^{-1}\sum_{c'\in G} Y_{c'}$ and
$\tilde S^{(G)}_{cj} = S_{cj} - |G|^{-1}\sum_{c'\in G} S_{c'j}$, 
and test, analogously to \eqref{equ:hypothesis_ci} and \eqref{eq:H0crp},
\begin{align*}
  H_{0,C,G}&:\ \tilde Y^{(G)} \perp\!\!\!\perp \tilde S_C \mid \tilde S_{\bar C} && \text{(activation)},\\
  H_{0,C,G}&:\ \tilde Y^{(G)}_c \perp\!\!\!\perp \tilde S^{(G)}_{cC} \mid \tilde S^{(G)}_{c\bar C} \quad \forall c \in G && \text{(CRP)},
\end{align*}
where $\tilde Y^{(G)}=(\tilde Y^{(G)}_c)_{c\in G}$.
At the root of $\mathcal T_G$, these are hypotheses~\eqref{equ:hypothesis_ci} and~\eqref{eq:H0crp}; for $G=\{a,b\}$, $\tilde Y^{(G)}_a = -\tilde Y^{(G)}_b = (Y_a-Y_b)/2$. 
As in Subsection~\ref{subsec:hierarchical_testing}, we call the pair $(C,G)$ null if $H_{0,C,G}$ holds and non-null otherwise.
A rejection of $H_{0,C,G}$ means that the units in $C$ carry information about differences between the classes in $G$ beyond the rest of the layer.

\begin{figure}[t!]
    \centering
    \includegraphics[width=0.7\linewidth]{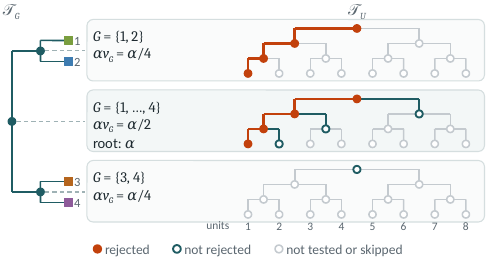}
    \caption{
    Top-down testing over the unit tree $\mathcal T_U$ and the class tree $\mathcal T_G$ for $K=4$ classes and $M_l=8$ units, with one run over $\mathcal T_U$ per internal node $G$ of $\mathcal T_G$. In the run for $G$, the pair $(C,G)$ is rejected if the parent of $C$ was rejected in this run and $\min(1, p_C M_l/|C|) \le \alpha v_G$. The pair is skipped if $(C,\mathrm{pa}(G))$ was not rejected, so the runs for $\{1,2\}$ and $\{3,4\}$ only test clusters rejected in the run for the root (middle). The weights are $v_G = 1/2$ at the root of $\mathcal T_G$ and $v_G = 1/4$ at the other two internal nodes, and the global root is tested at level $\alpha$ (Appendix~\ref{app:fwer}). The FWER over all $(C,G)$ is at most $\alpha$.
  In this example, unit 1 is rejected for the differences among all four classes and also for the difference between classes 1 and 2, whereas no cluster is rejected for the difference between classes 3 and 4.}
    \label{fig:node_tests}
\end{figure}

We fix weights $v_G \ge 0$ with $\sum_G v_G \le 1$ in advance and, for each $G$, apply the procedure of Subsection~\ref{subsec:hierarchical_testing} on $\mathcal{T}_U$ to $H_{0,C,G}$, $C \in \mathcal{T}_U$, at level $\alpha v_G$, except that we test the global root, i.e.\ the root of $\mathcal{T}_U$ paired with the root of $\mathcal{T}_G$, at level $\alpha$. We use $v_{\mathrm{root}} = 1/2$ and $v_G = 1/(2(K-2))$ for the other internal nodes, so that the weights sum to one, but any other choice fixed in advance is valid; for $K = 2$, the root is the only internal node and $v_{\mathrm{root}} = 1$. To reduce the number of tests, we process the nodes $G$ top-down and, for a non-root $G$, skip $(C,G)$ if $(C,\operatorname{pa}(G))$ was not rejected, where $\operatorname{pa}(G)$ denotes the parent of $G$ in $\mathcal{T}_G$. Figure~\ref{fig:node_tests} illustrates the procedure. The procedure controls the FWER over all pairs $(C,G)$ with $C \in \mathcal{T}_U$ and $G$ an internal node of $\mathcal{T}_G$ at level $\alpha$ (Appendix~\ref{app:fwer}). Skipping preserves this control, but for CRP scores it can cost power (Appendix~\ref{app:nesting}).

To test several layers $l = 1, \dots, L$, we apply the procedure to layer $l$ at level $\alpha w_l$, with weights $w_l \ge 0$, $\sum_l w_l \le 1$, fixed in advance, e.g.\ $w_l = 1/L$. This controls the FWER over all layers at level $\alpha$, whereas testing each layer at level $\alpha$ controls it only at level $L\alpha$ (Appendix~\ref{app:fwer}).

\subsection{Interpretation}\label{subsec:interpretation_framework}

A rejection is a population-level statement about the fixed network, the layer and $P_{XY}$; a non-rejection is not evidence that $(C,G)$ is null, since the tests may lack power (Subsection~\ref{subsec:tests}). We state the interpretation for $H_{0,C}$; for $H_{0,C,G}$, the response differences are those between the classes in $G$.
For activations, a rejection of $H_{0,C}$ means that the units in $C$ carry information about the response differences beyond the rest of the layer, whether or not the network uses them. For CRP scores, it means that the network uses the units in $C$ differently across classes, and that this difference carries information about the response differences beyond the relevance of the rest of the layer. Ranking units by relevance, in contrast, also selects units that the network uses but whose scores carry no information (Section~\ref{sec:sim}).
If $(C,G)$ is rejected but $(C',G)$ for none of the children $C'$ of $C$, the children may lack power or be null although $C$ is non-null, e.g.\ if the units in $C$ are strongly correlated because they detect the same or co-occurring patterns, so that the information of each child is contained in the units outside it. Similarly, if $(C,G)$ is rejected but $(C,G')$ for none of the children $G'$ of $G$, the information may concern differences between the two groups $G'$ rather than within them.

A rejection does not by itself imply that the patterns detected by the units in $C$ drive the response differences. Since $\tilde S_C$ is a function of $X$, it implies that the input carries information about these differences beyond the patterns detected by the other units. If the scores of the layer capture all information in the input about the differences, i.e.\ $\tilde Y \perp\!\!\!\perp X \mid \tilde S$, the converse also holds, so that $H_{0,C}$ is equivalent to $\tilde Y \perp\!\!\!\perp X \mid \tilde S_{\bar C}$; for CRP scores, this condition excludes additional information in units that the network does not use. That the patterns drive the differences further requires that no factor outside the input that affects the responses, such as a property of the genomic region, co-occurs with them.

\subsection{Conditional attribution maps}
\label{sec:attribution-maps}

For a rejected pair $(C, G)$, we visualize which input patterns the units in $C$ detect by conditional attribution maps \citep{CRP}. For an input $X$ and a class $c \in G$, we initialize the relevance at the output $f_{\hat\theta}(X)_c$, propagate it to layer $l$, set the relevance of all units outside $C$ to zero, and propagate the remaining relevance further to the input. The map assigns to each position and nucleotide the part of the prediction for class $c$ that passes through the units in $C$. 
To summarize the maps of a unit over inputs, we use relevance-weighted nucleotide frequencies in the simulation (Appendix~\ref{app:sim_detailed_results}) and TF-MoDISco tool (\citep{modisco}) in the application (Appendix~\ref{app:modisco-pipeline}).

The maps are descriptive. They are computed only for the selected clusters and summarize only the inputs with the largest relevance, so no error control is claimed for them. Whether the detected patterns drive the response differences requires the conditions in Subsection~\ref{subsec:interpretation_framework}.

\section{Simulation}\label{sec:sim}

We check FWER control and power where the ground truth is known. Details are given in Appendix~\ref{app:sim_details} and the simulation design is illustrated in Figure~\ref{fig:sim_design}. 
\subsection{Design}

\begin{figure}[t!]
    \centering
    \includegraphics[width=0.4\linewidth]{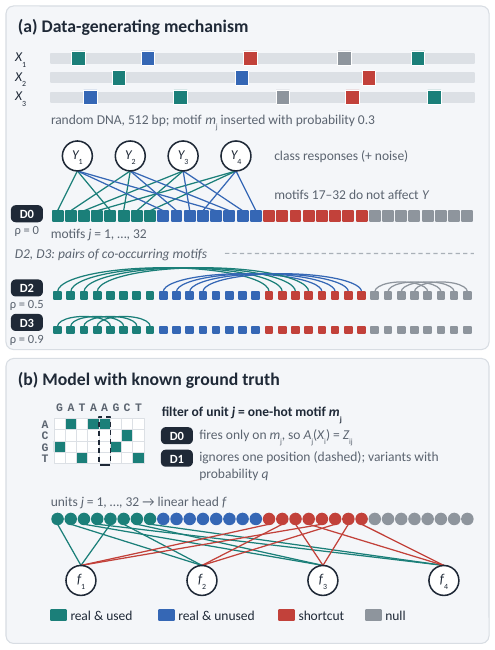}
    \caption{Simulation design. (a) Motif $m_j$ is inserted into $X_i$ if $Z_{ij}=1$; lines connect each motif to the class responses $Y_c$ it affects for $\delta=1$. In D2 and D3, arcs connect pairs of motifs with $\operatorname{Corr}(Z_{ij},Z_{ij'})=\rho$. (b) The filter of unit $j$ is the one-hot encoding of $m_j$; in D1, it ignores one position (dashed). Lines connect each unit to the outputs $f_c$ that use it.}
    \label{fig:sim_design}
\end{figure}

{\bf Data and network}. In our baseline scenario D0, we draw $n$ random DNA sequences $X_i$ of length 512 and $M = 32$ random 8-mers $m_1,\dots,m_M$, and insert motif $m_j$ into $X_i$ if $Z_{ij} = 1$, with $Z_{ij} \sim \mathrm{Ber}(0.3)$ independently. The responses of the $K = 4$ classes are
\begin{align*}
    Y_{ic} = \delta\,(Z_{ic} + Z_{i,c+4} + Z_{i,c+8} + Z_{i,c+12}) + \varepsilon_{ic},
\end{align*}
where $\varepsilon_{ic} \overset{\text{iid}}{\sim} N(0,1)$ independently of $X_i$, and $\delta\in \{0,1\}$. The tested layer has one convolutional unit per motif, whose filter is the one-hot encoding of $m_j$, so that unit $j$ detects motif $j$. A linear head maps the units to the outputs
\begin{align*}
    f_c(X) = 1 + \sum_{j \in U_c} \sum_t A_{jt}(X), 
\end{align*}
where $U_c = \{c,\, c+4,\, c+16,\, c+20\}$ and $A_{jt}(X)$ is the activation of unit $j$ at position $t$ (Subsection~\ref{subsec:unit_scores}).
Hence the motifs of units 1--16 affect $Y$ (real units) and the network uses units 1--8 and 17--24 (used units), which gives four groups of eight units: real \& used, real \& unused, shortcut, and null.
Shortcut units mimic units a trained network relies on although their scores carry no information about $Y$ beyond the real units \citep{geirhos2020shortcut}, e.g.\ because the training sample was not drawn from $P_{XY}$.

\textbf{Scenario variations.} Three variations of D0 mimic properties of trained networks (Appendix~\ref{app:scenarios}): filters that match their motif only approximately (D1), so that with probability $q \in \{0.1, 0.25\}$ unit $j$ misses an occurrence of $m_j$ or is activated by a variant that does not affect $Y$; shortcuts whose motifs co-occur with real ones (D2); and redundant units that detect co-occurring motifs (D3). In D2 and D3, pairs of motifs have $\operatorname{Corr}(Z_{ij}, Z_{ij'}) = \rho$ (Figure~\ref{fig:sim_design}a). D2 ($\rho = 0.5$) pairs the shortcuts 17--20 with the real \& used units 1--4 and the shortcuts 21--24 with the real \& unused units 9--12, and the null units 25--28 with 29--32. D3 ($\rho = 0.9$) pairs units 1--4 with 5--8.

{\bf Ground truth.} Let $\mathcal A$ be the set of units whose scores carry information about $\tilde Y$ given the scores of all other units. In all scenarios, $C$ is non-null iff $C \cap \mathcal A \neq \emptyset$ (Appendices~\ref{app:known_truth} and~\ref{app:scenarios}). For $\delta = 0$, $\mathcal A = \emptyset$. For $\delta = 1$, $\mathcal A$ contains the real units for activations and the real \& used units for CRP scores, and, for CRP scores in D2, also the shortcuts 21--24, as proxies of units 9--12, which are removed before testing.

{\bf Procedure and baselines.} For $n \in \{200, 400, 800, 2000, 4000\}$ and 200 replications, we build the tree on one half of the data (Subsection~\ref{subsec:clustering}) and traverse it top-down on the other half with the max GCM at $\alpha = 0.05$ (Subsection~\ref{subsec:tests}), only for the root of $\mathcal T_G$, i.e.\ hypotheses~\eqref{equ:hypothesis_ci} and~\eqref{eq:H0crp}, separately for activation and CRP scores, estimating the conditional means by post-lasso regressions with the R package \texttt{comets} \citep{kook24comets}. We compare with three baselines that consider only single units. The \emph{flat conditional} and the \emph{marginal} test apply the max GCM to each single unit $j$, 
given all other units and given no units, respectively, and reject if $M_l\, p_{\{j\}} \le \alpha$ (Bonferroni). \emph{Top-$k$}, without error control, selects on the inference sample the $k$ units with the largest mean absolute score, summed over classes for CRP scores, with $k = 16$ for activation and $k = 8$ for CRP scores. Except for CRP scores in D2, where $|\mathcal{A}| = 12$, this gives $k = |\mathcal{A}|$ for $\delta = 1$, which favours top-$k$.

{\bf Metrics.} The FWER is the proportion of replications with at least one false rejection. For $\delta = 0$, it equals the rejection rate of the root; for $\delta = 1$, it also counts false rejections below the root. Following \citet{meinshausen2008hierarchical}, the power for a class $I$ of cluster sizes is the proportion of non-null clusters $C$ with $|C| \in I$ that are rejected, pooled over replications (Appendix~\ref{app:metrics}). For top-$k$, we treat selecting unit $j$ as rejecting $H_{0,\{j\}}$, so its FWER is the proportion of replications in which it selects a unit outside $\mathcal A$, and its power is that for single units.

\subsection{Results}\label{subsec:results}

Figures~\ref{fig:fwer}--\ref{fig:leave_rej_d2d3} in Appendix~\ref{app:sim_details} show all results.
In all scenarios, for both scores and both values of $\delta$, the hierarchical max GCM keeps the FWER close to $\alpha$ (Monte Carlo standard error $\approx 0.015$).
In D0, its power increases with $n$ and with the cluster size, and approaches 1 for $n \ge 2000$. 
It is lowest for single units, which are tested at level $\alpha/M_l$ and only after all their ancestors are rejected.
With activations, the procedure detects the real \& used and the real \& unused units, and with CRP scores only the real \& used units.
Top-$k$ selects a unit outside $\mathcal{A}$ in almost every replication, so its FWER is close to 1.
Its power stays near $1/2$ for all $n$, since the mean absolute scores do not depend on $Y$ and are exchangeable across the units in the tree.
The conditional attribution maps of the units rejected for CRP scores in D0 recover the motifs of the real \& used units, whereas top-$k$ also selects shortcuts with equally clean motifs (Figure~\ref{fig:crp_selection_motifs}).
Since all motifs are random 8-mers, the maps alone cannot separate shortcuts from real \& used units; the test does.

In D1, power decreases with $q$, since the imperfect filters reduce the information that the scores carry about the motifs; for $q = 0.25$, it approaches 1 only for $n \ge 2000$.
In D2, the marginal test rejects the shortcuts correlated with real units, and its FWER approaches 1, whereas the hierarchical max GCM keeps it close to $\alpha$.
For CRP scores and $n \ge 2000$, the hierarchical max GCM also rejects the shortcuts 21--24, which are in $\mathcal{A}$ as proxies of the real \& unused units 9--12. As discussed in Subsection~\ref{subsec:interpretation_framework}, such a rejection does not imply that their motifs drive $Y$.
In D3, each unit of a pair carries little information beyond the other, so single units are reliably detected only for $n \ge 2000$ (activations) or $n = 4000$ (CRP).
Since the hierarchical and the flat conditional procedure both test single units at level $\alpha/M_l$, the hierarchy cannot increase their power. It localizes the information to the pair instead: it rejects a cluster within almost every pair for all $n$, whereas the flat conditional test needs $n \ge 2000$.

\section{Application}\label{sec:application}

We apply the procedure to a convolutional sequence-to-activity (S2A) network trained to predict DNA activity across cell types. CRP scores every unit and input window but does not say which units matter; instead of ranking hundreds of correlated units by relevance, we test, with error control, whether the relevance of a unit or cluster of units carries information about the cell-type responses beyond the other units of the layer. A rejected unit is thus cell-type-specific and non-redundant, and the class tree $\mathcal T_G$ ties it to the cell types it is relevant for.

\subsection{Data, model, and interpretability target}

The CNN maps a 512-bp one-hot-encoded zebrafish DNA sequence to pseudobulk ATAC-seq accessibility in 12 cell types, as a multivariate regression with one output per cell type (Appendix~\ref{app:appendix-bioapplication}). We apply the procedure to CRP relevance scores of the units in each convolutional layer, computed as described in Appendix~\ref{app:crp-bio}. The natural concept for such a model is a transcription factor (TF) motif, a short DNA sequence that a TF binds to regulate nearby genes. Curated resources provide position weight matrices for numerous TFs and annotate which TFs are active in which cell types \citep{JASPAR}, so we expect the patterns of selected units to match the motifs of TFs active in the corresponding cell types (Appendix~\ref{app:modisco-pipeline}). For evaluation, we use a reference set of 18{,}000 held-out windows, the 1{,}500 most cell-type-specific windows for each of the 12 cell types (Appendix~\ref{app:appendix-bioapplication}).

\subsection{Results}

\begin{figure*}[t]
  \centering
  \includegraphics[width=\textwidth]{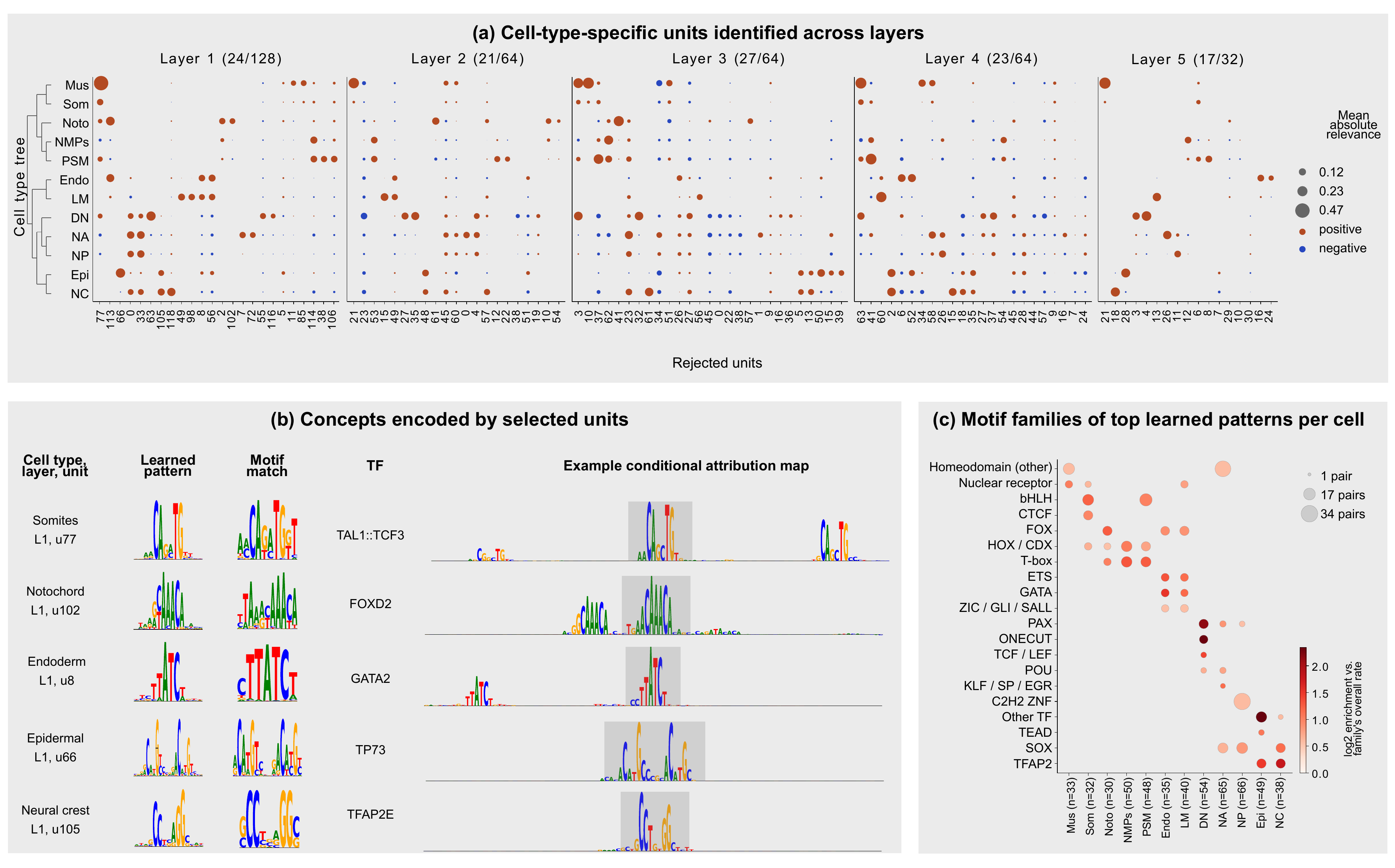}
    \caption{Hierarchical testing localizes cell-type-specific concepts across layers. 
    (a) Units selected per layer, with selected/total units in parentheses. Each layer is tested at level $0.005$ (Subsection~\ref{subsec:localizing_classes}), so the FWER over all five layers is at most $0.025$.
    Rows are cell types ordered by lineage tree; dot size shows mean absolute relevance and color the contribution sign. (b) Example first-layer units with best-matching motif and CRP attribution map. (c) Motif families matched by each cell type's top patterns; dot size is the number of unit-motif pairs (n per cell type) and color is log2 enrichment relative to the family's overall rate.}
    \label{fig:bio_application}
\end{figure*}

Applied to the trained S2A model, our method selects 112 of 352 units across the five convolutional layers (Fig.~\ref{fig:bio_application}a), narrowing the network to a small set of units that can be inspected individually. First-layer units are mostly specific to a single cell type, whereas units in layers 2--4 are often shared among cell types that are adjacent on the cell type tree (e.g., the neural group), reflecting the lineage structure of the data.

We matched a learned pattern for each selected unit, derived from its conditional attribution maps, to the core JASPAR motif database \citep{JASPAR}. First-layer units match TFs with known roles in the corresponding cell types, for example GATA2 for Endoderm, TFAP2E for Neural crest and TP73 for Epidermal, and the attribution maps place the pattern at the matched position (Fig.~\ref{fig:bio_application}b). Across cell types, the enriched motif families follow the cell type tree (Fig.~\ref{fig:bio_application}c): SOX in neural cell types, HOX/CDX and T-box in posterior mesoderm, and GATA, ETS and FOX in Endoderm and LM. As no motif information was given to the model during training, this agreement with known biology provides an independent check that the selected units capture meaningful concepts.

\section{Conclusion}

We proposed a hierarchical testing procedure to localize predictive information in the representations of a neural network with error control. Within a layer of a fixed trained network, it identifies the units and clusters of units whose scores carry information about the response differences between classes beyond the rest of the layer. The procedure controls the FWER over the whole hierarchy. 
Applying the procedure to activations and to CRP scores separates two questions: which units encode information about the responses, and which of these the network relies on. 
In the simulation, our procedure controlled the FWER in all scenarios and, with CRP scores, detected the units that are both informative and used by the network. Unlike testing single units conditionally with Bonferroni correction, it localized the information to clusters when correlated units could not be separated individually, whereas testing them marginally or ranking them by relevance also selected shortcuts that carry no information beyond the real units.

In the application to a sequence-to-activity model of zebrafish DNA accessibility, the procedure selected a set of units small enough to inspect one by one, whose patterns are consistent with known transcription factor motifs. The error control applies to the selection of units; the attribution maps and the motif matching are descriptive.

A rejection in our procedure is a statement about the fixed network, the layer and the distribution of the inference sample. It does not by itself imply that the detected input patterns drive the responses (Subsection~\ref{subsec:interpretation_framework}). Power depends on how well the GCM regressions estimate the conditional means and on the tree, and strongly correlated units may only be detected jointly as a cluster. 
We evaluated the procedure only on DNA sequence models and for differences between class responses, although it applies to any input space and to responses that are not compared across classes (Section~\ref{sec:hierarchical-testing}).
The most natural extension is to test all layers jointly with error control that exploits their dependence, rather than a union bound over layers. Further directions are controlling the FDR instead of the FWER, and replacing units by other components of a representation, such as directions in activation space.

\subsubsection*{Acknowledgments}
M. Simnacher, J. Berend, W. Samek, C. Lippert, S. Greven and U. Ohler were supported by DFG research unit DeSBI (KI-FOR 5363). V. Huryn was supported by a graduate fellowship in the Helmholtz International Berlin Research School for Data Science (HEIBRiDS). 
\subsection*{AI use statement}

In this work, we used generative AI tools for help develop theoretical models or conceptual frameworks;  refine hypotheses; design or provide feedback on research methodology or experiments; implement methods; interpret results. We have not used generative AI tools to generate synthetic data sets; clean and reformat datasets; support qualitative and thematic data analysis; discover research topics or identify gaps. Additionally, we used generative AI tools for create or modify scientific figures or images; suggest experimental parameters; create or edit software code; draft parts of a research paper; summarize or analyze existing literature; identify relevant literature;. We have reviewed all AI-assisted work by proofreading and editing generated code, text, and images. We have personally read and verified every paper cited in this work. We take responsibility for the final content of this work, including text, claims, or artifacts produced with the aid of generative AI. 



\bibliographystyle{unsrtnat}
\bibliography{references}  

\appendix
\input{appendices}

\end{document}

%% file: math_commands.tex
\usepackage{amsmath,amsfonts,bm}

\def\eqref#1{equation~\ref{#1}}

\def\1{\bm{1}}

\DeclareMathAlphabet{\mathsfit}{\encodingdefault}{\sfdefault}{m}{sl}
\SetMathAlphabet{\mathsfit}{bold}{\encodingdefault}{\sfdefault}{bx}{n}



%% file: appendices.tex
\section*{Appendix} 
\section{Method details}\label{app:method}

\subsection{Correlation of CRP scores}\label{app:crp_corr}

For CRP scores ($H = K$), we correlate the scores over all inputs and classes after subtracting from $\tilde S_{hj}$ its mean over inputs for each class $h$,
\[
\rho_{jk} = \frac{\sum_{h=1}^K \operatorname{Cov}(\tilde S_{hj}, \tilde S_{hk})}{\sqrt{\sum_{h=1}^K \operatorname{Var}(\tilde S_{hj}) \sum_{h=1}^K \operatorname{Var}(\tilde S_{hk})}},
\]
with (co)variances over the inputs of the clustering sample. Without subtracting the means over inputs, units with similar mean relevance per class would be correlated even if their scores vary independently across inputs, although such units are not redundant for the tests. For $K = 2$, $\rho_{jk}$ reduces to the correlation of the class contrasts $R_j(X,a) - R_j(X,b)$ and $R_k(X,a) - R_k(X,b)$.

\subsection{Monte Carlo p-value of the max GCM}\label{app:pvalue}

Let $\Xi^{(1)},\dots,\Xi^{(B)}$ be independent draws from $N(0,\hat\Gamma)$, with entries $\Xi^{(b)}_{cj}$ indexed like those of $T_i$. The p-value of node $C$ is
\[
p_C = \frac{1}{B+1}\Big(1 + \sum_{b=1}^{B} \mathds 1\Big\{\max_{c\in \{1,\ldots,K\},\,j \in C} |\Xi^{(b)}_{cj}| \ge W^{\max}_C\Big\}\Big)
\]
\citep[Section~3.2]{shah2020hardness}. Since $\sum_c \tilde Y_c = 0$, the entries of $T_i$ can be linearly dependent, e.g.\ for activations, where all classes are regressed on the same scores, so $\hat\Gamma$ can be singular. This affects neither the draws nor $W^{\max}_C$, since neither requires inverting $\hat\Gamma$.

\subsection{FWER control over the unit and class trees}\label{app:fwer}

All probabilities are conditional on the training and clustering samples, so both trees are fixed. For fixed $G$, the run over $\mathcal{T}_U$ is the procedure of Subsection~\ref{subsec:hierarchical_testing} at level $\alpha v_G$, which controls the FWER over $H_{0,C,G}$, $C \in \mathcal{T}_U$, at level $\alpha v_G$ if the p-values are valid \citep{meinshausen2008hierarchical, goeman2010sequential}. Skipping only adds a condition for rejection and can therefore only remove rejections. Let $C_0$ and $G_0$ be the roots of $\mathcal{T}_U$ and $\mathcal{T}_G$. If $(C_0,G_0)$ is non-null, rejecting it is not an error, and the bound for the run for $G_0$ is unchanged by testing it at level $\alpha$, since the bound depends only on the levels at which null pairs are tested. By the union bound over $G$, the FWER is then at most $\sum_G \alpha v_G \le \alpha$. If $(C_0,G_0)$ is null, every other pair is tested only after $(C_0,G_0)$ is rejected: within a run, $(C,G)$ is tested only after $(C_0,G)$, and for $G \ne G_0$, $(C_0,G)$ only after $(C_0,\operatorname{pa}(G))$. The FWER is then at most the probability of rejecting $(C_0,G_0)$, i.e.\ at most $\alpha$. Neither case uses relations between the hypotheses, so the result holds for activation and CRP scores alike, and asymptotically for asymptotically valid p-values. For several layers tested at levels $\alpha w_l$, the union bound over layers gives FWER control at level $\sum_l \alpha w_l \le \alpha$, and at level $L\alpha$ if each layer is tested at level $\alpha$.

\subsection{Nesting of the hypotheses and power}\label{app:nesting}

Nesting affects only power. By the weak union property of conditional independence, $H_{0,C,G}$ implies $H_{0,C',G}$ for $C' \subset C$, so the ancestors in $\mathcal{T}_U$ of a non-null pair are non-null. For activations, the hypotheses are also nested in $G$, since $\tilde Y^{(G')}$ is a function of $\tilde Y^{(G)}$ for $G' \subset G$, so skipping never discards a non-null pair. For CRP scores, which are also centered within $G$, $(C,G')$ can be non-null although $(C,\operatorname{pa}(G'))$ is null, and skipping it then costs power.

\section{Simulation details}
\label{app:sim_details}

\subsection{Data-generating mechanism}\label{app:dgm}
We draw sequences of length $p = 512$ with nucleotides drawn uniformly from $\{A, C, G, T\}$, one-hot encoded as $X_i \in \{0,1\}^{4 \times p}$, and $M = 32$ random 8-mers $m_1, \dots, m_M$. If $Z_{ij} = 1$, motif $m_j$ is inserted into $X_i$ at a randomly drawn position that does not overlap other inserted motifs. If a sequence contains a motif elsewhere by chance, it is redrawn. The responses are generated as in Section~\ref{sec:sim}.

\subsection{Network and ground truth}\label{app:known_truth}
We describe the network and the ground truth for D0; Appendix~\ref{app:scenarios} gives the changes for D1–D3.
We specify a CNN with a single convolutional layer of $M$ units, ReLU activation, global average pooling, and a linear head. The filter of unit $j$ has width 8, no padding, bias $-7$, and weights given by the one-hot encoding of motif $m_j$. Writing $L_t \in \{A,C,G,T\}$ for the letter at position $t$ of $X$, unit $j$ computes, for $t = 1, \dots, p-7$,
\[
A_{jt}(X) = \max\Big(0,\ \sum_{s=1}^{8} \mathbf 1\{L_{t+s-1} = m_{js}\} - 7\Big).
\]
The sum counts the positions in the window starting at $t$ that agree with $m_j$, so $A_{jt}(X) = 1$ if the window equals $m_j$ and $0$ otherwise. Since sequences containing a motif by chance are redrawn, motif $j$ occurs in $X_i$ exactly once if $Z_{ij} = 1$ and not at all otherwise. Hence the activation score of Subsection~\ref{subsec:unit_scores} is $A_j(X_i) = \max_t A_{jt}(X_i) = Z_{ij}$, i.e. unit $j$ indicates exactly whether motif $j$ occurs in $X_i$, and global average pooling outputs $Z_{ij}/(p-7)$ for unit $j$. The linear head has weight $p-7$ from each unit $j \in U_c = \{c, c+4, 16+c, 20+c\}$ to output $c$, zero weights otherwise, and bias $1$, so that
\[
f_c(X_i) = 1 + \sum_{j \in U_c} Z_{ij}.
\]

Since the head is linear, the relevance of unit $j$ for class $c$, summed over positions, equals its contribution to $f_c$, $R_j(X, c) = \mathds 1\{j \in U_c\}\, \sum_t A_{jt}(X)$, up to a factor $\kappa \approx 0.9995$ in our implementation (Appendix~\ref{app:implementation}).

For $\delta = 1$ and activations, $\tilde Y$ depends on $X$ only through $Z_{i1},\dots,Z_{i16}$, the scores of units 1--16, each with a nonzero coefficient, and the $Z_{ij}$ are independent. Hence, given $\tilde S_{\bar C}$, $\tilde Y$ depends on $\tilde S_C$ iff $C$ contains one of units 1--16, so $\mathcal A = \{1,\dots,16\}$. For CRP scores, units 9--16 and 25--32 have $R_j \equiv 0$ and are removed before testing (Subsection~\ref{subsec:hierarchical_testing}), so the CRP-based trees contain the 16 used units, with centered scores $\tilde S_{cj} = (\mathds 1\{j \in U_c\} - 1/K)\,Z_{ij}$. The part of $\tilde Y_c$ not determined by these scores depends only on $Z_{i9},\dots,Z_{i16}$ and $\varepsilon_i$, which are independent of the scores of all used units. Hence, given $\tilde S_{c\bar C}$, $\tilde Y_c$ depends on $\tilde S_{cC}$ iff $C$ contains one of units 1--8, so $\mathcal A = \{1,\dots,8\}$. In both cases, $H_{0,C}$ is false iff $C \cap \mathcal A \neq \emptyset$. For $\delta = 0$, $\tilde Y$ is independent of $X$, so every $H_{0,C}$ holds.

\subsection{Scenario variations}\label{app:scenarios}

Scenarios D1--D3 modify the data-generating mechanism of Appendix~\ref{app:dgm} and, in D1, the filters of Appendix~\ref{app:known_truth}. The responses $Y_{ic}$, the linear head and the procedure are as in D0, and indicators $Z_{ij}$ not mentioned below are independent as in D0.

\paragraph{D1: imperfect filters.}
For each unit $j$, we draw a position $u_j$ uniformly from $\{1, \dots, 8\}$ once for each value of $q$, the same for all $n$, $\delta$ and replications, and set the filter weights at position $u_j$ to zero and the bias to $-6$, so that
\[
  A_{jt}(X) = \max\Big(0,\ \sum_{s \neq u_j}
    \mathds{1}\{L_{t+s-1} = m_{js}\} - 6\Big)
\]
equals one iff the window starting at $t$ agrees with $m_j$ at all positions except $u_j$. For $s \in \{1, \dots, 8\}$, a variant of $m_j$ in position $s$ is $m_j$ with its letter at position $s$ replaced by one of the three other letters, drawn uniformly. Let $D_{ij}, E_{ij} \sim \mathrm{Ber}(q)$ be independent of each other and of all other variables. If $Z_{ij} = 1$ and $D_{ij} = 1$, we insert, instead of $m_j$, a variant of $m_j$ in a position drawn uniformly from $\{1, \dots, 8\} \setminus \{u_j\}$. If $E_{ij} = 1$, we additionally insert a variant of $m_j$ in position $u_j$. The responses depend on $Z_{ij}$ as in D0, so the first variant affects $Y$ as $m_j$ does, whereas the second does not affect $Y$. Sequences in which a filter fires at a window not covered by an inserted motif or variant are redrawn. Hence
\[
  A_j(X_i) = \max\{Z_{ij}(1 - D_{ij}),\ E_{ij}\},
\]
with $P(A_j(X_i) = 0 \mid Z_{ij} = 1) = q(1-q)$ and $P(A_j(X_i) = 1 \mid Z_{ij} = 0) = q$.

\paragraph{D2 and D3: correlated units.}
For a pair $(j, j')$ of units, $Z_{ij}$ is drawn as in D0 and
\begin{align*}
  Z_{ij'} &= V_{ij'} Z_{ij} + (1 - V_{ij'}) Z'_{ij'}, \\
  V_{ij'} &\sim \mathrm{Ber}(\rho), \quad Z'_{ij'} \sim \mathrm{Ber}(0.3),
\end{align*}
with $V_{ij'}$ and $Z'_{ij'}$ independent of each other and of all other variables. Then $Z_{ij'} \sim \mathrm{Ber}(0.3)$ as in D0 and $\mathrm{Cov}(Z_{ij}, Z_{ij'}) = \rho \cdot 0.3 \cdot 0.7$, so $\mathrm{Corr}(Z_{ij}, Z_{ij'}) = \rho$. D2 ($\rho = 0.5$) pairs units 1--4 with 17--20, 9--12 with 21--24, and 25--28 with 29--32, each in order (e.g., 1 with 17), and D3 ($\rho = 0.9$) pairs units 1--4 with 5--8. 

\paragraph{Ground truth.}
In D1, the CRP scores of unit $j$ sum the relevance over positions and are therefore proportional to the number of windows at which unit $j$ fires, $N_{ij} = Z_{ij}(1 - D_{ij}) + E_{ij}$. Both $A_j(X_i)$ and $N_{ij}$ are functions of $(Z_{ij}, D_{ij}, E_{ij})$, which are independent across $j$, and both carry information about $Z_{ij}$, since $P(A_j(X_i) = 1 \mid Z_{ij} = 1) - P(A_j(X_i) = 1 \mid Z_{ij} = 0) = (1-q)^2 > 0$ and $E[N_{ij} \mid Z_{ij} = 1] - E[N_{ij} \mid Z_{ij} = 0] = 1 - q > 0$.
Hence, given the scores of all other units in the tree, the scores of unit $j$ carry information about $\tilde Y$ iff $Z_{ij}$ enters $\tilde Y$, i.e.\ $\mathcal{A} = \{1, \dots, 16\}$ for activations and $\mathcal{A} = \{1, \dots, 8\}$ for CRP scores.

In D2 and D3, $A_j(X_i) = Z_{ij}$ as in D0, but the indicators of a pair are dependent. Since $V_{ij'}$ and $Z'_{ij'}$ are independent of all other variables, a unit $j'$ whose indicator does not enter $\tilde Y$, i.e.\ a shortcut or null unit in D2, carries no information about $\tilde Y$ beyond its partner $j$ if $j$ is in the tree. A real unit carries information beyond its partner, since its indicator enters $\tilde Y$ and $\rho < 1$. For activations, all 32 units are in the tree, so $\mathcal{A} = \{1, \dots, 16\}$. For CRP scores, units 9--16 and 25--32 have $R_j \equiv 0$ and are removed before testing. In D2, the partners 9--12 of the shortcuts 21--24 are therefore not in the tree. Since $Z_{i,c+8}$ enters $\tilde Y_c$ and is correlated with $Z_{i,20+c}$, the scores of unit $20+c$ carry information about $\tilde Y_c$ given the scores of all other used units. Hence $\mathcal{A} = \{1, \dots, 8, 21, \dots, 24\}$ in D2 and $\mathcal{A} = \{1, \dots, 8\}$ in D3. In all scenarios, given $\tilde S_{\mathcal A}$, the scores of the units outside $\mathcal A$ depend only on variables that are independent of $(\tilde Y, \tilde S_{\mathcal A})$: their own $(Z_{ij}, D_{ij}, E_{ij})$, which do not enter $\tilde Y$, and, in D2, $(V_{ij'}, Z'_{ij'})$ and the indicator of the partner, which either does not enter $\tilde Y$ or is determined by $\tilde S_{\mathcal A}$. Hence $\tilde Y \perp\!\!\!\perp \tilde S_{\bar{\mathcal A}} \mid \tilde S_{\mathcal A}$, where $\bar{\mathcal A}$ denotes the units in the tree outside $\mathcal A$ (for CRP scores, for the scores of each class $c$), and by weak union $H_{0,C}$ holds iff $C \cap \mathcal A = \emptyset$.

\subsection{Performance measures}\label{app:metrics}
Let $\mathcal R_l$ be the set of rejected clusters and $\mathcal T_{U,l}$ the unit tree in replication $l = 1,\dots,200$. For top-$k$, $\mathcal R_l = \{\{j\} : j \text{ selected}\}$. With $\mathcal D_l(I) = \{C \in \mathcal T_{U,l} : |C| \in I,\ C \cap \mathcal A \neq \emptyset\}$, the non-null clusters of size in $I$,
\begin{align*}
    \widehat{\mathrm{FWER}} &= \frac{1}{200}\sum_{l=1}^{200} \mathds 1\{\exists\, C \in \mathcal R_l : C \cap \mathcal A = \emptyset\}, \\
    \widehat{\mathrm{Power}}(I) &= \frac{\sum_{l} |\mathcal D_l(I) \cap \mathcal R_l|}{\sum_{l} |\mathcal D_l(I)|}.
\end{align*}
Since top-$k$ selects only single units, we report its power only for $I = \{1\}$.

\subsection{Implementation}\label{app:implementation}
At node $C$, we use $B = \max\{999, \lceil 20 M_l/(\alpha|C|)\rceil\}$ draws in the p-value of Appendix~\ref{app:pvalue}, so that about 20 draws fall into the rejection region at the level $\alpha|C|/M_l$ at which $C$ is tested (Subsection~\ref{subsec:hierarchical_testing}). This gives $B = 999$ at the root and 12,800 (activations, $M_l=32$) or 6,400 (CRP, $M_l=16$) at the leaves.
The lasso penalty is chosen by cross-validation ($\lambda_{\min}$ of \texttt{cv.glmnet} \citep{glmnet}), and the residuals are those of least-squares refits on the selected scores.

CRP scores are computed with the code of the application, which uses zennit \citep{anders2026software} and its CRP extension \citep{CRP}. We apply LRP-0 \citep{bach2015pixel, montavon2019layer}, i.e.\ zennit's $\varepsilon$-rule with $\varepsilon = 0$, and sum the relevance of each unit over positions. The stabilizer of zennit's rule for average pooling scales all relevances by $\kappa = (1/505)/(1/505 + 10^{-6}) \approx 0.9995$, so $R_j(X, c) = \kappa\, \mathds 1\{j \in U_c\}\, \sum_t A_{jt}(X)$. Since the tree and the tests are invariant to rescaling the scores, $\kappa$ does not affect the results.

\subsection{Detailed results}\label{app:sim_detailed_results}

\begin{figure*}[t]
    \centering
    \includegraphics[width=\linewidth]{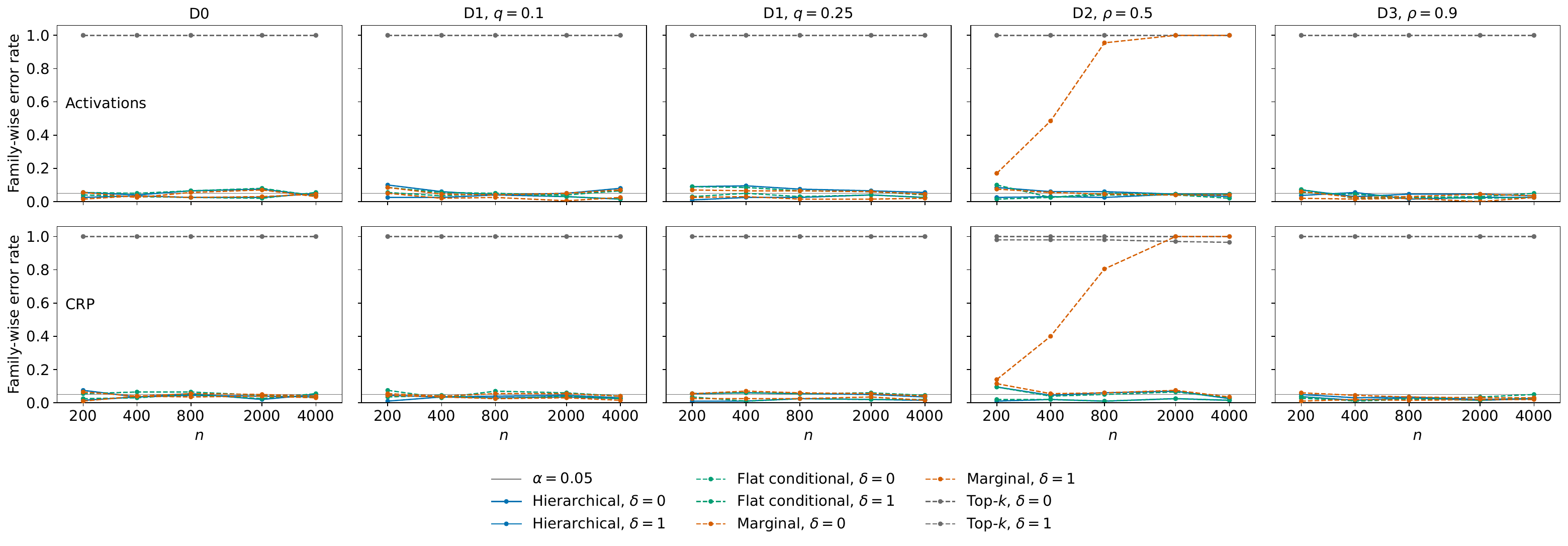}
    \caption{FWER of the hierarchical max GCM and the three baselines for activation scores (top) and CRP scores (bottom) in scenarios D0--D3 (columns), each estimated from 200 replications. Line style and colour indicate the method (legend); within each method, the lighter shade is $\delta = 0$ and the darker shade $\delta = 1$. For $\delta = 0$, all  units are null, so top-$k$ has FWER 1 by construction. The horizontal line marks $\alpha = 0.05$.}
    \label{fig:fwer}
\end{figure*}

\begin{figure*}[t]
    \centering
    \includegraphics[width=\linewidth]{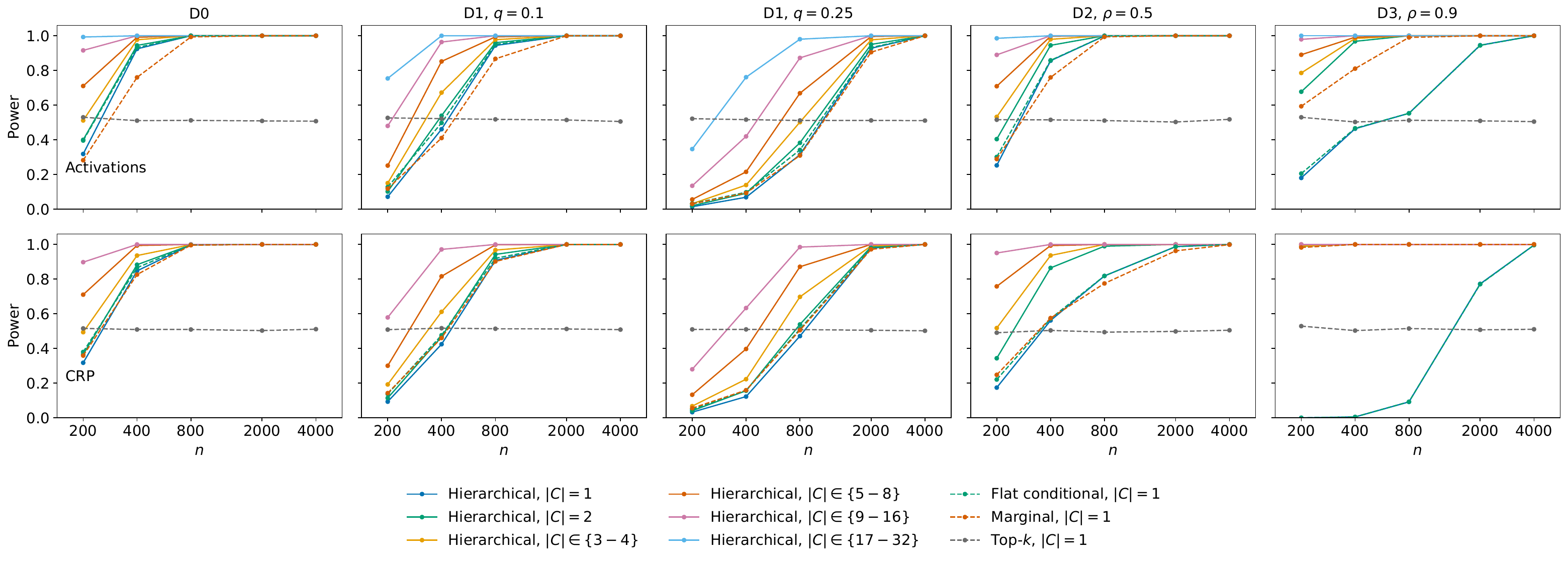}
    \caption{Power for $\delta = 1$ for activation scores (top) and CRP scores (bottom) in scenarios D0--D3 (columns), each estimated from 200 replications. For the hierarchical max GCM, each curve is the proportion of non-null clusters $C$ with $|C| \in I$ that are rejected, for $I \in \{\{1\}, \{2\}, \{3, 4\}, \{5, \dots, 8\}, \{9, \dots, 16\},
\{17, \dots, 32\}\}$. For the three baselines, it is the proportion for single units. Since unused units are removed before testing CRP scores, the CRP trees contain 16 units, and $I = \{17, \dots, 32\}$ is shown only for activations.}
    \label{fig:power}
\end{figure*}

\begin{figure}[t]
    \centering
    \includegraphics[width=\linewidth]{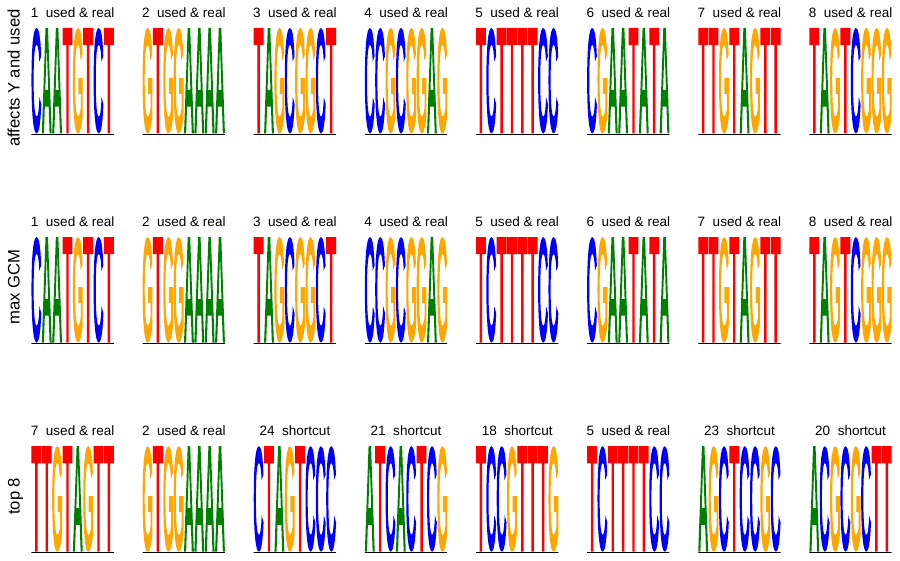}
    \caption{Motifs recovered in one replication ($n = 2000$, $\delta = 1$, in D0). Each panel shows the relevance-weighted nucleotide frequencies of one unit (Subsection~\ref{sec:attribution-maps}), titled by the unit and its group. Top: the true motifs $m_1,\dots,m_8$ of the real \& used units. Middle: the units rejected by the hierarchical max GCM on CRP scores, here units 1--8, each recovering its motif. Bottom: the eight units selected by top-$k$, five of which are shortcuts.}
    \label{fig:crp_selection_motifs}
\end{figure}

\begin{figure}[t]
    \centering
    \includegraphics[width=\linewidth]{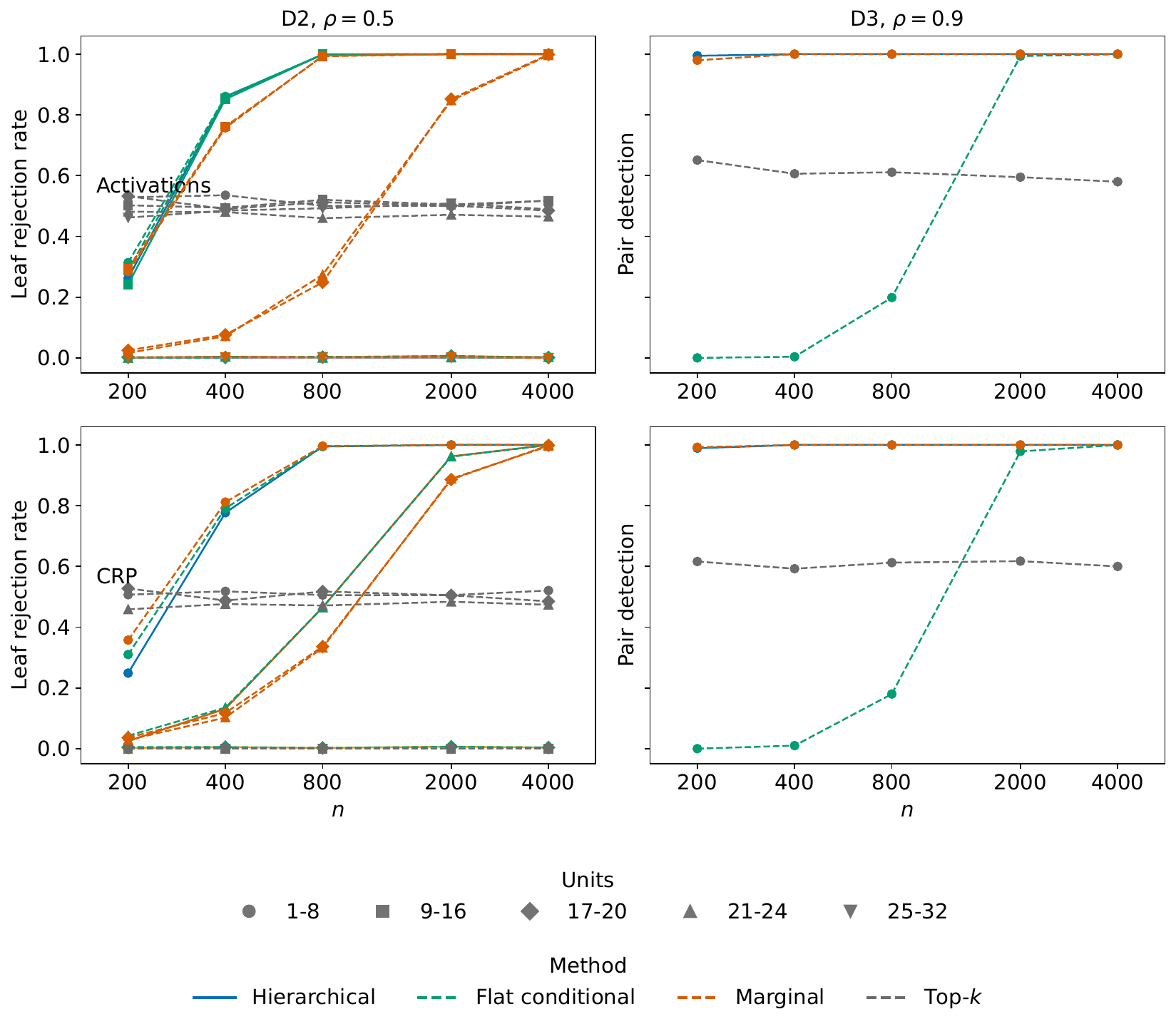}
    \caption{Leaf rejection in D2 (left) and pair detection in D3 (right) for
$\delta = 1$, for activation scores (top) and CRP scores (bottom), each
estimated from 200 replications. Left: proportion of replications and units
of a group (marker) for which the single unit $\{j\}$ is rejected, or
selected by top-$k$. Units 1--4 are paired with the shortcuts 17--20, units
9--12 with the shortcuts 21--24, and units 25--28 with 29--32. Rejections of
units 17--20 and 25--32 are false, and so are rejections of units 21--24 for
activation scores. For CRP scores, units 9--16 and 25--32 are removed before
testing. Right: proportion of replications and of the four redundant pairs
$\{j, j+4\}$, $j = 1, \dots, 4$, for which a rejected cluster lies within the
pair, i.e.\ the pair or one of its units is rejected, or, for top-$k$, a unit
of the pair is selected.}
    \label{fig:leave_rej_d2d3}
\end{figure}

Figures~\ref{fig:fwer}, \ref{fig:power} and \ref{fig:leave_rej_d2d3} show the FWER and the power for all $n$ and cluster sizes, each estimated from 200 replications. 
To summarize the conditional attribution maps (Subsection~\ref{sec:attribution-maps}) of a unit $j$ over inputs, we align the input windows with the largest relevance in the map of $\{j\}$ and compute the relevance-weighted nucleotide frequencies per position. For the first-layer units of the simulation, these approximate the motif each filter matches. Figure~\ref{fig:crp_selection_motifs} shows the motifs recovered by the conditional attribution maps (Subsection~\ref{sec:attribution-maps}) in one replication of D0 with $n = 2000$, $\delta = 1$ and CRP scores (Section~\ref{subsec:results}).

\section{Additional tests at each node}
\label{app:tests}

In addition to the max GCM from Subsection~\ref{subsec:tests}, we provide alternative CITs for our setting below.
All results in Sections~\ref{sec:sim} and~\ref{sec:application} use the max GCM; we include these tests as alternatives within the same procedure but do not evaluate them here.
At a node $(C, G)$ of Subsection~\ref{subsec:localizing_classes}, each test is applied to the responses $\tilde Y^{(G)}_c$, $c \in G$, and, for CRP scores, to the scores $\tilde S^{(G)}$, with $K$ replaced by $|G|$.

\subsection{Quadratic GCM}

Alternatively, the residual products of Subsection~\ref{subsec:tests} can be aggregated by $W_C = n\,\bar T^\top \hat\Sigma^{-1} \bar T$, where $\hat\Sigma$ is the empirical covariance matrix of $T_1,\dots,T_n$, with p-value from the $\chi^2_{q_C}$ distribution \citep{kook24comets}. Since $\hat\Sigma$ must be invertible, we drop the entries of class $K$, so that $q_C = (K-1)|C|$. For activations, $W_C$ then does not depend on which class is dropped if the regressions are linear in the response. For CRP scores, the test has no power against violations of \eqref{eq:H0crp} only in class $K$; with a linear network after layer $l$, the entries of class $K$ are linear combinations of the others, and nothing is lost. Inverting $\hat\Sigma$ is unstable if $q_C$ is large relative to $n$, which can inflate the type I error.

\subsection{Wald-type test}

\paragraph{Linear model.}
We restrict to a linear model for the conditional mean of the responses. Writing \(\tilde S_{ihC}\in\mathbb{R}^{|C|}\) for the \(h\)-th row of \(\tilde S_{iC}\), we assume for $c=1,\ldots,K-1$
\begin{equation}\label{eq:linear_model}
\tilde Y_{ic}=\beta_{0c}
+\sum_{h=1}^H B_{ch,C}^\top\,\tilde S_{ihC}
+\sum_{h=1}^H B_{ch,\bar C}^\top\,\tilde S_{ih\bar C}
+\varepsilon_{ic},
\end{equation}
with \(\varepsilon_i=(\varepsilon_{i1},\ldots,\varepsilon_{i,K-1})^\top\), \(\mathbb{E}[\varepsilon_i\mid\tilde S_i]=0\) and \(\mathrm{Cov}(\varepsilon_i)=\Omega\in\mathbb{R}^{(K-1)\times(K-1)}\). Under \(H_{0,C}\), the conditional mean of \(\tilde Y\) does not depend on \(\tilde S_C\), so all coefficients \(B_{ch,C}\) of the units in \(C\) vanish, which is the hypothesis we test. The intercepts \(\beta_{0c}\) are not attributed to any unit.

\emph{Activation scores} (\(H=1\)). With \(A_{iC}=\tilde S_{i1C}\), the model has class-specific coefficients,
\[
\tilde Y_{ic}=\beta_{0c}+\beta_{c,C}^\top A_{iC}+\beta_{c,\bar C}^\top A_{i\bar C}+\varepsilon_{ic},
\]
and the null hypothesis considered instead of the null hypothesis in \eqref{equ:hypothesis_ci} is
\begin{align*}
    H_{0,C}^{\mathrm{lin}}:\ \beta_{1,C}=\ldots=\beta_{K-1,C}=0.
\end{align*}

\emph{CRP scores} (\(H=K\)). The response of class \(c\) is regressed on the relevance for class \(c\) only, with coefficients common to all classes, i.e.\ \(B_{ch,C}=\mathds{1}\{h=c\}\,\beta_C\) and
\(B_{ch,\bar C}=\mathds{1}\{h=c\}\,\beta_{\bar C}\) in \eqref{eq:linear_model},
so that the sum over \(h\) reduces to the term \(h=c\),
\[
\tilde Y_{ic}=\beta_{0c}+\beta_C^\top\tilde S_{icC}+\beta_{\bar C}^\top\tilde S_{ic\bar C}+\varepsilon_{ic},
\]
with considered null hypothesis $H_{0,C}^{\mathrm{lin}}:\ \beta_C=0$.
For \(K=2\), both models reduce to a regression of the response contrast on the activations or on the relevance contrasts \(R_j(X,a)-R_j(X,b)\), respectively.

\paragraph{Estimation and test.}
We write both models as \(\tilde Y_i=Z_i\theta+\varepsilon_i\), with
\begin{align*}
Z_i&=I_{K-1}\otimes\big(1,\ A_{i\bar C}^\top,\ A_{iC}^\top\big)
\ \ \text{(activations)},\\
Z_{i}&=\big[\,I_{K-1}\ \ \tilde S_{i\bar C}^{[K-1]}\ \ \tilde S_{iC}^{[K-1]}\,\big]
\ \ \text{(CRP)},
\end{align*}
where \(\tilde S^{[K-1]}\) omits row \(K\), and write \(\theta_C\) for the
\(q_C\) coefficients of the units in \(C\) (\(q_C=(K-1)|C|\) for activations,
\(q_C=|C|\) for CRP). We estimate \(\theta\) by feasible generalized least
squares,
\[
\hat\theta=Q^{-1}\sum_i Z_i^\top\hat\Omega^{-1}\tilde Y_i,\qquad
Q=\sum_i Z_i^\top\hat\Omega^{-1}Z_i,
\]
where \(\hat\Omega\) is the empirical covariance of pooled least-squares
residuals. For activations, this coincides with equation-wise least squares.
With the cluster-robust covariance
\(\hat V=Q^{-1}\big(\sum_i Z_i^\top\hat\Omega^{-1}\hat\varepsilon_i\hat\varepsilon_i^\top\hat\Omega^{-1}Z_i\big)Q^{-1}\),
the Wald statistic \(W_C=\hat\theta_C^\top\hat V_{CC}^{-1}\hat\theta_C\) is
asymptotically \(\chi^2_{q_C}\)-distributed under \(H_{0,C}^{\mathrm{lin}}\)
and yields the \(p\)-value \(p_C\) used in
Subsection~\ref{subsec:hierarchical_testing}.

\section{Zebrafish DNA Accessibility Model}
\label{app:appendix-bioapplication}

\subsection{Data}

We used a published dataset on DNA accessibility \citep{Zebrahub} that profiled by single-cell ATAC-seq across a zebrafish
developmental time course spanning six stages (0--30 somites), yielding 94{,}562 cells. Cells were pseudobulked into 12 embryonic cell types used for
model training: \textit{Differentiating Neurons, Endoderm, Epidermal, Lateral Mesoderm, Muscle, Neuromesodermal progenitors (NMPs), Neural Anterior, Neural Crest, Neural Posterior, Notochord, Presomitic Mesoderm (PSM),} and \textit{Somites}. An additional \textit{Adaxial Cells} pseudobulk (390 cells) was excluded for insufficient depth, and an \textit{unassigned} population was excluded throughout.

We used such cell type abbreviations in figures: Mus, muscle; Som, somites; Noto, notochord; NMPs, neuromesodermal progenitors; PSM, presomitic mesoderm; Endo, endoderm; LM, lateral mesoderm; DN, differentiating neurons; NA/NP, neural anterior/posterior; Epi, epidermal; NC, neural crest.

Consensus accessible regions were called from the pooled Tn5 insertion data, giving 640{,}834 peaks (median width 576\,bp) against the \texttt{danRer11} zebrafish genome assembly. Fixed-width 512\,bp windows were tiled around each peak summit without overlaps, each window was assigned to its nearest summit. This yielded 887{,}999 labelled windows. For each window and cell type, the regression target is the raw Tn5 insertion count summed over the window, normalized by that cell type's genome-wide sequencing depth (counts per million) and $\log(1+x)$-transformed.

Cell-type specificity of each sequence $i$ was quantified with the $\tau$ index
\citep{Tau_index},
\[
\tau_i = \frac{1}{K-1}\sum_{c=1}^{K}\left(1 - \bar{Y}_{i,c}\right), \qquad
\bar{Y}_{i,c} = \frac{Y_{i,c}}{\max_{j} Y_{i,j}},
\] where $Y_{i,c}$ is the CPM-normalized signal (linear scale) for sequence $i$ in cell type $c$, $\bar{Y}_{i,c}$ is $Y_{i,c}$ divided by the maximum signal across cell types for that sequence, $j \in \{1,\dots,K\}$ indexes the cell types over which the maximum is taken, and $K=12$ is the number of cell types. $\tau_i$ ranges from 0 for sequences with equal signal in all cell types to 1 for sequences with signal in a single cell type. Sequences with zero signal in all cell types were excluded.

Sequences were one-hot encoded ($A,C,G,T \to \mathbb{R}^4$; ambiguous bases mapped to the zero vector) and extracted from the reference FASTA at training time. During training, each window was reverse-complemented with probability 0.5 (accessibility is strand-agnostic, so the target is unchanged); validation windows were always read on the forward strand. Chromosomes 1--21 and 24--25 were used for training and chromosomes 22--23 for validation, split at the chromosome level to prevent sequence leakage between windows derived from the same genomic region. 

For method evaluation, we assembled a reference set of $n=18{,}000$ windows restricted to the held-out validation chromosomes (22--23). These chromosomes were not used for gradient updates, but they were used to select the best checkpoint based on validation metrics. Each window $i$ was assigned to the cell type $c^\ast_i=\arg\max_c Y_{i,c}$ with the highest measured target signal. Within each cell type $c$, windows with $c^\ast_i = c$ were ranked by the cell-type specificity index $\tau$ (higher = more specific), and the top 1{,}500 were kept. The union over the $K=12$ cell types yields the final set of 18{,}000 windows. Therefore, the set is class-balanced and enriched for cell-type-specific sequences.

\subsection{Model Architecture}

Sequence-to-activity prediction uses a compact, fully convolutional CNN chosen to stay directly compatible with layer-wise relevance propagation under zennit/CRP: no max-pooling and no flatten+dense readout, since both break clean relevance conservation. The network has five 1-D convolutional blocks over the one-hot input ($4 \times 512$), each consisting of a convolution, ReLU, and (except the final block) $2\times$ average pooling:

\begin{center}
\begin{tabular}{ccccc}
\hline
Layer & Channels & Kernel & Pool & $\lambda$ \\
\hline
1 & 128 & 21 & $\times2$ & 0.0 \\
2 & 64  & 5  & $\times2$ & 0.01 \\
3 & 64  & 3  & $\times2$ & 0.01 \\
4 & 64  & 5  & $\times2$ & 0.01 \\
5 & 32  & 5  & --        & 0.01\\
\hline
\end{tabular}
\end{center}

No batch normalization is used (found empirically to distort CRP relevance attribution); an $\ell_1$ penalty ($\lambda=0.01$) is instead applied to the post-activation feature maps of layers 2--5 to encourage a sparse code at the depth where combinatorial, cell-type-specific concepts are expected to emerge. The resulting receptive field is 148\,bp (29\% of the 512\,bp input). Global average pooling over the positional axis of the final feature map (32 channels) yields a 32-dimensional vector. A single linear layer maps this vector to 12 outputs, one per cell type. The full model has 95{,}468 trainable parameters.

\subsection{Training Procedure}

The loss combines three terms on the log1p-transformed targets $Y$ and predictions $\hat{Y}$, each window-and-cell-type vector of length 12:
\begin{align*}
\mathcal{L} &= w_{\text{mse}}\,\mathrm{MSE}(\hat{Y}, Y)
+ w_{\text{cos}}\big(1 - \cos(\hat{Y}, Y)\big) \\
&\quad + w_{\text{spec}}\,\mathrm{MSE}(\hat{Y}_c, Y_c),
\end{align*}

where $\hat{Y}_c, Y_c$ are each window's prediction/target after subtracting its own cross-cell-type mean. The MSE term rewards correct absolute accessibility magnitude, the cosine term rewards correct relative shape of the 12-cell-type profile, and the centered-MSE term isolates the cell-type-\emph{specificity} residual directly, since the first two terms are dominated by the $\sim$94\% of signal shared across cell types (mean-predictor Pearson $r = 0.94$, specificity $r \approx 0$). Entries with $y < 1$ (log1p scale; accessibility near the detection floor) receive $2\times$ the squared-error weight in the MSE term, since such entries are rare but otherwise contribute negligible gradient.

Training used a two-stage curriculum, both stages with the same low-target upweighting ($y<1 \Rightarrow 2\times$ squared-error weight) and reverse-complement augmentation described above. (1) \emph{Pretraining} on all 887{,}999 windows regardless of specificity with the MSE term weighted most heavily ($w_{\text{mse}}=10$, $w_{\text{cos}}=1$, $w_{\text{spec}}=1$) so the model first learns overall accessibility, using AdamW with lr $=10^{-3}$ and effectively no weight decay ($10^{-20}$). (2) \emph{Fine-tuning} of the full network initialized from the stage-1 checkpoint (no layers frozen) on the 419{,}572 windows with the high specificity ($\tau > 0.52$), with the loss re-weighted toward shape and specificity ($w_{\text{mse}}=0.5$, $w_{\text{cos}}=10$, $w_{\text{spec}}=5$) using AdamW with lr $=3\times10^{-3}$ and weight decay $10^{-4}$. Both stages used cosine-annealing learning-rate decay and batch size 256.

On the held-out validation chromosomes (22--23, 26{,}174 windows), the fine-tuned model reaches a mean per-cell-type Pearson correlation of 0.62 and a mean cell-type-\emph{specificity} Pearson correlation (computed on cell-type-centered predictions/targets) of 0.55 — exceeding a $k$-mer-count ridge-regression baseline ($r=0.44$ raw, $r=0.25$ specificity) on both axes.

\subsection{Concept relevance propagation and conditional attribution maps}
\label{app:crp-bio}

\paragraph{Relevance propagation.}
We explained the trained CNN with Concept Relevance Propagation (CRP) \citep{CRP}, using \texttt{zennit} v0.4.6 and \texttt{zennit-crp} v0.6.0. CRP extends layer-wise relevance propagation (LRP) by restricting the backward pass to a chosen subset of channels (``concepts'') of an intermediate layer, which yields relevance maps conditional on those concepts. Each convolutional filter (unit) is treated as one concept: the channel axis of each \texttt{Conv1d} output is the concept axis, and relevance is summed over positions when a scalar per unit is needed.

The first convolutional layer, which acts on the one-hot input, used the Flat rule. All subsequent convolutional layers and the linear head used LRP-0 ($\varepsilon=0$), which is fully conservative and has no free hyperparameter. All attributions were computed with the model in evaluation mode.

\paragraph{Class--unit relevance.}
For a cell type $c$, relevance was initialised at the corresponding output $f_c(X)$ and propagated back to the output of convolutional layer $l$. For a window $X$, this gives the relevance of unit $u$,
\[
R^{(l)}_{u\leftarrow c}(X)=\sum_{t} r^{(l)}_{u,t}(X),
\]
where $t$ indexes positions. Relevances are signed: positive values mean that unit $u$ pushes the prediction for cell type $c$ up, negative values that it pushes it down.

\paragraph{Conditional attribution maps.}
To obtain base-resolution maps for an individual unit, we used CRP conditioning: for window $X$, layer $l$ and unit $u$, all other channels of layer $l$ were masked during the backward pass, so that relevance reaches the input only through unit $u$. With relevance seeded at the output for cell type $c$, the resulting \emph{class-conditional map} shows which bases feed unit $u$ on its way to cell type $c$. Computing such maps for every $c$ on a shared set of windows allows the same unit to be compared across cell types. Where a map should describe what a unit detects irrespective of cell type, we instead seeded relevance at the raw activation of unit $u$ (an \emph{activation-seeded map}), so that nothing downstream of the unit enters the map. The motif analysis uses class-conditional maps. Because the input is one-hot, we take the relevance on the observed nucleotide at each position as the per-base contribution and set it to zero on the other three nucleotides.

\section{Assigning TF motifs to units}
\label{app:modisco-pipeline}

The class-tree test selects units that are specific to a cell type, but it does not say what a unit encodes. We assign motifs to each selected (unit, cell type) pair $(u,c)$, called a \emph{pair} below, in four steps (Fig.~\ref{fig:modisco_scheme}). A cell type is assigned to a unit only if it lies within one of the unit's narrowest rejected nodes of the cell-type tree; a rejection of the root node alone is not used. A unit assigned to several cell types is analysed once per cell type, and each analysis is independent.

\paragraph{1. Window pool.}
For a pair $(u,c)$, we start from the reference windows with dominant cell type $c$ (Appendix~\ref{app:appendix-bioapplication}). We rank them by the signed CRP relevance $R^{(l)}_{u\leftarrow c}(X)$ of unit $u$ for cell type $c$ and keep the top 500.

\paragraph{2. Conditional attribution.}
For each kept window, we compute the class-conditional map of unit $u$ for cell type $c$ (Appendix~\ref{app:crp-bio}). This gives one attribution map of shape $4\times512$ (nucleotide $\times$ position) per window. It applies to all layers, so units in deeper layers are also described through their effect on the input sequence.

\paragraph{3. Pattern discovery.}
We run TF-MoDISco-lite v2.4.0 \citep{modisco} on the one-hot sequences and the per-base contributions of each pair. It extracts seqlets (window 15\,bp, flank 5\,bp) from the attributions, clusters them, and reports consolidated patterns (FDR 0.05). Each pattern is a contribution weight matrix (CWM) with a seqlet count, reported separately for positive and negative attribution. We cap the number of seqlets per metacluster at 20{,}000. For display, a CWM is trimmed to the positions whose summed absolute contribution is at least 30\% of the maximum, plus 4\,bp of padding on each side. This is the ``learned pattern'' in Fig.~\ref{fig:bio_application}b.

\paragraph{4. Motif matching.}
We match each pattern to the JASPAR CORE vertebrate 2025 collection; 1{,}019 motifs) with the TOMTOM-lite matcher implemented in TF-MoDISco-lite, which ranks database motifs by Euclidean distance between position-weight matrices. A match with lowest p-value is reported.

\begin{figure*}[t]
    \centering
    \includegraphics[width=\linewidth]{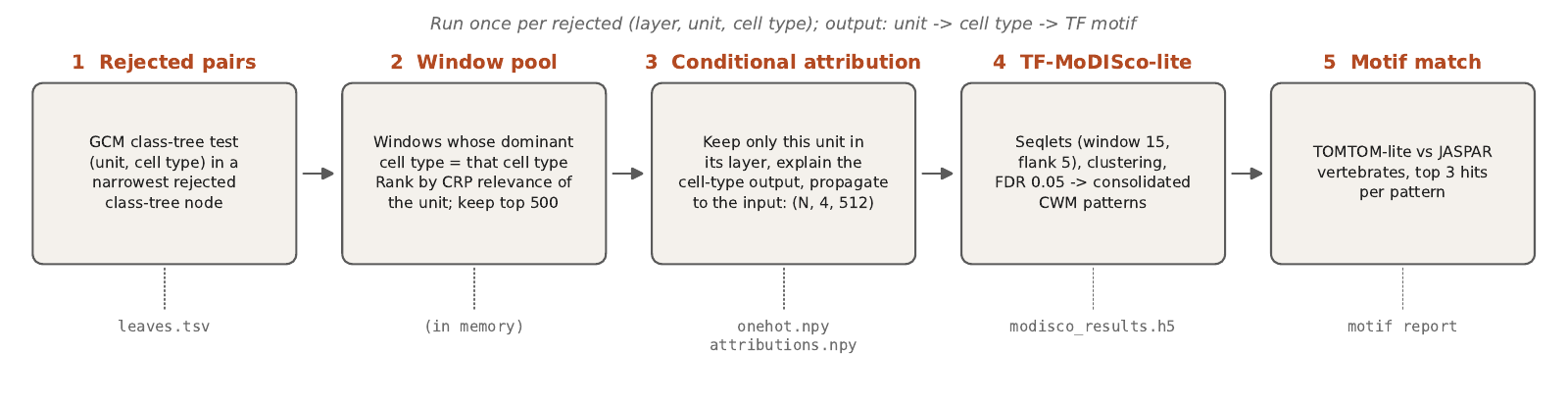}
    \caption{Motif to unit matching pipeline.}
    \label{fig:modisco_scheme}
\end{figure*}

%% file: references.bib
@article{CRP,
  title = {From attribution maps to human-understandable explanations through Concept Relevance Propagation},
  volume = {5},
  ISSN = {2522-5839},
  url = {http://dx.doi.org/10.1038/s42256-023-00711-8},
  DOI = {10.1038/s42256-023-00711-8},
  number = {9},
  journal = {Nature Machine Intelligence},
  publisher = {Springer Science and Business Media LLC},
  author = {Achtibat,  Reduan and Dreyer,  Maximilian and Eisenbraun,  Ilona and Bosse,  Sebastian and Wiegand,  Thomas and Samek,  Wojciech and Lapuschkin,  Sebastian},
  year = {2023},
  month = Sept,
  pages = {1006–1019}
}

@article{bach2015pixel,
  title={On pixel-wise explanations for non-linear classifier decisions by layer-wise relevance propagation},
  author={Bach, Sebastian and Binder, Alexander and Montavon, Gr{\'e}goire and Klauschen, Frederick and M{\"u}ller, Klaus-Robert and Samek, Wojciech},
  journal={PloS one},
  volume={10},
  number={7},
  pages={e0130140},
  year={2015},
  publisher={Public Library of Science San Francisco, CA USA}
}

@Article{glmnet,
    title = {Regularization Paths for Generalized Linear Models via
      Coordinate Descent},
    author = {Jerome Friedman and Trevor Hastie and Robert Tibshirani},
    journal = {Journal of Statistical Software},
    year = {2010},
    volume = {33},
    number = {1},
    pages = {1--22},
    doi = {10.18637/jss.v033.i01},
  }

@article{montavon2019layer,
  title={Layer-wise relevance propagation: an overview},
  author={Montavon, Gr{\'e}goire and Binder, Alexander and Lapuschkin, Sebastian and Samek, Wojciech and M{\"u}ller, Klaus-Robert},
  journal={Explainable AI: interpreting, explaining and visualizing deep learning},
  pages={193--209},
  year={2019},
  publisher={Springer}
}

@article{anders2026software,
  title={Software for dataset-wide XAI: from local explanations to global insights with Zennit, CoRelAy, and ViRelAy},
  author={Anders, Christopher J and Neumann, David and Samek, Wojciech and M{\"u}ller, Klaus-Robert and Lapuschkin, Sebastian},
  journal={PloS one},
  volume={21},
  number={1},
  pages={e0336683},
  year={2026},
  publisher={Public Library of Science San Francisco, CA USA}
}

@article{CRAFT,
  doi = {10.48550/ARXIV.2211.10154},
  url = {https://arxiv.org/abs/2211.10154},
  author = {Fel,  Thomas and Picard,  Agustin and Bethune,  Louis and Boissin,  Thibaut and Vigouroux,  David and Colin,  Julien and Cadène,  Rémi and Serre,  Thomas},
  title = {CRAFT: Concept Recursive Activation FacTorization for Explainability},
  publisher = {arXiv},
  year = {2022},
  copyright = {Creative Commons Attribution 4.0 International}
}

@misc{ACE,
  doi = {10.48550/ARXIV.1902.03129},
  url = {https://arxiv.org/abs/1902.03129},
  author = {Ghorbani,  Amirata and Wexler,  James and Zou,  James and Kim,  Been},
  title = {Towards Automatic Concept-based Explanations},
  publisher = {arXiv},
  year = {2019},
  copyright = {arXiv.org perpetual,  non-exclusive license}
}

@article{bykov2023labeling,
  title={Labeling neural representations with inverse recognition},
  author={Bykov, Kirill and Kopf, Laura and Nakajima, Shinichi and Kloft, Marius and H{\"o}hne, Marina},
  journal={Advances in Neural Information Processing Systems},
  volume={36},
  pages={24804--24828},
  year={2023}
}

@article{teneggi2024testing,
  title={Testing semantic importance via betting},
  author={Teneggi, Jacopo and Sulam, Jeremias},
  journal={Advances in Neural Information Processing Systems},
  volume={37},
  pages={76450--76499},
  year={2024}
}

@article{TCAV,
  doi = {10.48550/ARXIV.1711.11279},
  url = {https://arxiv.org/abs/1711.11279},
  author = {Kim,  Been and Wattenberg,  Martin and Gilmer,  Justin and Cai,  Carrie and Wexler,  James and Viegas,  Fernanda and Sayres,  Rory},
  title = {Interpretability Beyond Feature Attribution: Quantitative Testing with Concept Activation Vectors (TCAV)},
  publisher = {arXiv},
  year = {2017},
  copyright = {arXiv.org perpetual,  non-exclusive license}
}

@article{Zebrahub,
  title = {Zebrahub-Multiome: Uncovering Gene Regulatory Network Dynamics During Zebrafish Embryogenesis},
  url = {http://dx.doi.org/10.1101/2024.10.18.618987},
  DOI = {10.1101/2024.10.18.618987},
  publisher = {openRxiv},
  author = {Kim,  Yang Joon and Vijay Kumar,  Shruthi and Iovino,  Benjamin and Granados,  Alejandro A. and Ancheta,  Sarah and Frank,  Max and Zhao,  Xiang and Awayan,  Kyle and Seng,  Amanda and Borja,  Michael and Paul,  Sheryl and Mekonen,  Honey and Arjyal,  Ritwicq and Detweiler,  Angela and Şenbabaoğlu,  Yasin and Gómez-Sj\"{o}berg,  Rafael and Neff,  Norma and Lange,  Merlin and Royer,  L\"{o}ıc A.},
  year = {2024},
  month = Oct 
}

@article{meinshausen2008hierarchical,
  title={Hierarchical testing of variable importance},
  author={Meinshausen, Nicolai},
  journal={Biometrika},
  volume={95},
  number={2},
  pages={265--278},
  year={2008},
  publisher={Oxford University Press}
}

@article{mandozzi2016hierarchical,
  title={Hierarchical testing in the high-dimensional setting with correlated variables},
  author={Mandozzi, Jacopo and B{\"u}hlmann, Peter},
  journal={Journal of the American Statistical Association},
  volume={111},
  number={513},
  pages={331--343},
  year={2016},
  publisher={Taylor \& Francis}
}

@article{JASPAR,
  title = {JASPAR 2026: expansion of transcription factor binding profiles and integration of deep learning models},
  volume = {54},
  ISSN = {1362-4962},
  url = {http://dx.doi.org/10.1093/nar/gkaf1209},
  DOI = {10.1093/nar/gkaf1209},
  number = {D1},
  journal = {Nucleic Acids Research},
  publisher = {Oxford University Press (OUP)},
  author = {Ovek Baydar,  Damla and Rauluseviciute,  Ieva and Aronsen,  Dina R and Blanc-Mathieu,  Romain and Bonthuis,  Ine and de Beukelaer,  Herman and Ferenc,  Katalin and Jegou,  Alice and Kumar,  Vipin and Lemma,  Roza Berhanu and Lucas,  Jérémy and Pochon,  Mathis and Yun,  Chang M and Ramalingam,  Vivekanandan and Deshpande,  Salil Sanjay and Patel,  Aman and Marinov,  Georgi K and Wang,  Austin T and Aguirre,  Alejandro and Castro-Mondragon,  Jaime A and Baranasic,  Damir and Chèneby,  Jeanne and Gundersen,  Sveinung and Johansen,  Morten and Khan,  Aziz and Kuijjer,  Marieke L and Hovig,  Eivind and Lenhard,  Boris and Sandelin,  Albin and Vandepoele,  Klaas and Wasserman,  Wyeth W and Parcy,  Fran\c{c}ois and Kundaje,  Anshul and Mathelier,  Anthony},
  year = {2025},
  month = Dec,
  pages = {D184–D193}
}

@article{kook24comets,
title={{Algorithm-Agnostic Significance Testing in Supervised Learning With Multimodal Data}},
author={Lucas Kook and Anton Rask Lundborg},
year={2024},
journal={Briefings in Bioinformatics},
volume={25},
number={6},
doi={10.1093/bib/bbae475},
}

@article{goeman2010sequential,
  title={The sequential rejection principle of familywise error control},
  author={Goeman, Jelle J and Solari, Aldo},
  journal={The Annals of Statistics},
  pages={3782--3810},
  year={2010},
  publisher={JSTOR}
}

@article{Tau_index,
  title = {Genome-wide midrange transcription profiles reveal expression level relationships in human tissue specification},
  volume = {21},
  ISSN = {1367-4803},
  url = {http://dx.doi.org/10.1093/bioinformatics/bti042},
  DOI = {10.1093/bioinformatics/bti042},
  number = {5},
  journal = {Bioinformatics},
  publisher = {Oxford University Press (OUP)},
  author = {Yanai,  Itai and Benjamin,  Hila and Shmoish,  Michael and Chalifa-Caspi,  Vered and Shklar,  Maxim and Ophir,  Ron and Bar-Even,  Arren and Horn-Saban,  Shirley and Safran,  Marilyn and Domany,  Eytan and Lancet,  Doron and Shmueli,  Orit},
  year = {2004},
  month = Sept,
  pages = {650–659}
}

@Article{shah2020hardness,
  title={The hardness of conditional independence testing and the generalised covariance measure},
  author={Shah, R. and Peters, J.},
  journal={The Annals of Statistics},
  volume={48},
  number={3},
  pages={1514--1538},
  year={2020},
  publisher={Institute of Mathematical Statistics}
}

@inproceedings{ACD_new,
  title     = {Hierarchical Interpretations for Neural Network Predictions},
  author    = {Singh, Chandan and Murdoch, W. James and Yu, Bin},
  booktitle = {International Conference on Learning Representations (ICLR)},
  year      = {2019},
  url       = {https://openreview.net/forum?id=SkEqro0ctQ}
}

@article{geirhos2020shortcut,
  title   = {Shortcut learning in deep neural networks},
  author  = {Geirhos, Robert and Jacobsen, J{\"o}rn-Henrik and Michaelis, Claudio and Zemel, Richard and Brendel, Wieland and Bethge, Matthias and Wichmann, Felix A.},
  journal = {Nature Machine Intelligence},
  volume  = {2},
  number  = {11},
  pages   = {665--673},
  year    = {2020}
}

@misc{modisco,
  doi = {10.48550/ARXIV.1811.00416},
  url = {https://arxiv.org/abs/1811.00416},
  author = {Shrikumar,  Avanti and Tian,  Katherine and Avsec,  Žiga and Shcherbina,  Anna and Banerjee,  Abhimanyu and Sharmin,  Mahfuza and Nair,  Surag and Kundaje,  Anshul},
  title = {Technical Note on Transcription Factor Motif Discovery from Importance Scores (TF-MoDISco) version 0.5.6.5},
  publisher = {arXiv},
  year = {2018},
  copyright = {arXiv.org perpetual,  non-exclusive license}
}

@inproceedings{paillard2025hierarchical,
  title={Hierarchical Variable Importance with Statistical Control for Medical Data-Based Prediction},
  author={Paillard, Joseph and Collas, Antoine and Engemann, Denis A and Thirion, Bertrand},
  booktitle={International Conference on Information Processing in Medical Imaging},
  pages={79--93},
  year={2025},
  organization={Springer}
}

@article{simnacher2026embedded,
  title={Embedded Conditional Independence Tests for Large Language Model Generated Text with an Application to German Parliament Speeches},
  author={Simnacher, Marco and Keilbar, Georg and K{\"o}nig, Benjamin and Lippert, Christoph and Greven, Sonja},
  journal={arXiv preprint arXiv:2609.00946},
  year={2026}
}

@article{simnacher2026deep,
  title={Deep nonparametric conditional independence tests for images},
  author={Simnacher, Marco and Xu, Xiangnan and Park, Hani and Lippert, Christoph and Greven, Sonja},
  journal={Journal of Machine Learning Research},
  volume={27},
  number={96},
  pages={1--73},
  year={2026}
}
